\pdfoutput=1

\documentclass{article}

\usepackage{arxiv}
\usepackage{hyperref}       
\usepackage{booktabs}       
\usepackage{amsfonts}       
\usepackage{graphicx}

\usepackage{multirow}
\usepackage{multicol}
\usepackage{amsmath}
\usepackage{algorithm}
\usepackage{algorithmic}
\usepackage{subfig}

\newtheorem{theorem}{Theorem}[section]
\newtheorem{lemma}{Lemma}[section]
\newtheorem{remark}{Remark}[section]
\newtheorem{assumption}{Assumption}[section]
\newtheorem{definition}{Definition}[section]

\title{Learning to optimize by multi-gradient \\for multi-objective optimization}

\date{} 					

\author{ {Linxi Yang}\\
	School of Mathematical Sciences\\
	Sichuan University\\
Sichuan China, 610064 \\
	\texttt{leoyanglinxi@gmail.com} \\
\And
{Xinmin Yang}\\
	National Center for Applied Mathematics in Chongqing\\
	Chongqing Normal University\\
Chongqing China, 401331 \\
	\texttt{xmyang@cqnu.edu.cn} \\

	\And
	{Liping Tang}\thanks{Corresponding author}  \\
	National Center for Applied Mathematics in Chongqing\\
	Chongqing Normal University\\
Chongqing China, 401331 \\
	\texttt{tanglipings@163.com} \\
}

\renewcommand{\shorttitle}{\textit{arXiv} Template}

\hypersetup{
pdftitle={A template for the arxiv style},
pdfsubject={},
pdfauthor={Linxi Yang, Xinmin Yang, Liping Tang},
pdfkeywords={multi-objective optimization, learning to optimize, stochastic gradient methods, safeguard},
}

\begin{document}
\maketitle

\begin{abstract}
	The development of artificial intelligence (AI) for science has led to the emergence of learning-based research paradigms, necessitating a compelling reevaluation of the design of multi-objective optimization (MOO) methods. The new generation MOO methods should be rooted in automated learning rather than manual design.
In this paper, we introduce a new automatic learning paradigm for optimizing MOO problems, and propose a multi-gradient learning to optimize (ML2O) method, which automatically learns a generator (or mappings) from multiple gradients to update directions.
As a learning-based method, ML2O acquires knowledge of local landscapes by leveraging information from the current step and incorporates global experience extracted from historical iteration trajectory data.
By introducing a new guarding mechanism, we propose a guarded multi-gradient learning to optimize (GML2O) method, and prove that the iterative sequence generated by GML2O converges to a Pareto critical point.
The experimental results demonstrate that our learned optimizer outperforms hand-designed competitors on training multi-task learning (MTL) neural network.
\end{abstract}


\section{Introduction}
Multi-objective optimization (MOO) is a popular research topic in optimization, where multiple learning objectives are solved simultaneously.
In general, it is usually impossible to find a single solution  satisfying all objectives best at the same time, but a Pareto optimal set, where improvement in some objective function can only be achieved with the cost of an impairment in some other objectives.
MOO problems has gained wide attention in many real-world scenarios such as semantic segmentation \cite{long2015fully}, online advertising models \cite{lin2019pareto} and autonomous driving systems \cite{huang2019apolloscape,lu2019l3}.
The selection of an appropriate optimization method is crucial to effectively address the MOO problem. This is because different objectives may exhibit conflicting behaviors, resulting in gradients pointing in opposite directions. Additionally, the scales of these objective gradients can vary significantly, leading to a dominance of the largest gradient when simply adding them together.
To address the challenge of conflicting objectives, numerous approaches have been extensively investigated.

One of the methods is scalarization \cite{geoffrion1968proper, gass1955computational}, which obtains the optimal solution to a MOO problem by solving one or several parametrized single-objective optimization problems. These methods are commonly referred to as weighting methods, where nonnegative linear combinations of objective functions are minimized \cite{drummond2008choice,johannes1984scalarization}.
These parameters are not known in advance and requires the researcher to make choices based on their understanding of the objective interaction.
However, the effectiveness of this approach relies on the quality of the weight settings, and simple averages are only applicable in non-conflicting objectives with uniform scales.
Therefore, researchers subsequently proposed the adaptive scalarization technique \cite{kendall2018multi}, wherein the parameters of the scalarization are automatically determined during the course of the algorithm to ensure optimization quality.
This type of method is then no longer limited to designing the weights themselves but to designing the method of weight generation.

Meanwhile, non-scalarized MOO methods have also garnered significant attention in recent years. One prominent method is the multiple objective gradient manipulation (MOGM)
\cite{sener2018multi, yu2020gradient, chen2018gradnorm}, which adjusts multiple gradients to identify a shared update direction while ensuring all objectives descend simultaneously.
This class of methods can be generalized to a general framework , where the update direction is a weighted combination of multiple gradients \cite{zhou2022convergence}. Within this framework, MOGM methods can be perceived as a way of generating combinatorial weights, where the weights for each iteration step are computed based on information derived from the current step.
Considering that in practical applications, these full-gradient based methods above can bring unacceptable computational costs,  researchers have increasingly turned their attention to MOGM methods in the stochastic case as a more pragmatic and efficient approach.
Given that the stochastic gradient noise in multi-objective scenarios can potentially lead to an update direction completely opposite to the intended one, mitigating the impact of such noise becomes a central focus for these methods.
In order to alleviate the perturbation caused by gradient noise through leveraging historical information, there exist techniques that integrate momentum-like terms into the gradient \cite{fernando2023mitigating} and weighting parameter components \cite{zhou2022convergence} of these methods, respectively.

The MOO methods described above consist of two components. One component determines the update direction of the current step solely based on the instantaneous gradient of the ongoing iteration \cite{sener2018multi, yu2020gradient, chen2018gradnorm}, while the other component incorporates historical information regarding the trajectory of the iteration \cite{fernando2023mitigating, zhou2022convergence}.
In fact, these MOO methods can be viewed as a type of generator (or mapping) for weighted weights, and the direction of the most rapid descent of the weighted combination of the objective function is the direction of the update suggested by the current method.
The commonality among these generators lies in the fact that their generation strategies are manually pre-specified. Thus the efficacy of the update directions heavily relies on the researcher's priori knowledge about the problem.
This gives rise to two limitations: 1) When confronted with complex problems (e.g., non-convex \cite{pardalos2017non}, constrained MOO \cite{drummond2004projected}, etc.) or problems where the explicit form of the objective function cannot be articulated (e.g., multi-task learning (MTL) \cite{zhang2021survey}), the computation involved in designing suitable weights can become exceedingly challenging.
Researcher must take into account both the local and global landscape of the objective function during algorithm design, as failure to do so may result in iterations easily stuck in local minima and yielding undesirable outcomes.
Meanwhile, manually designed methods often contain undetermined hyperparameters, researchers often resort to a trial and error approach to sift through the myriad of potential parameter combinations and identify the optimal solution for the given problem at hand. However, this process is undoubtedly time-consuming and inefficient.
2) The generalizability of algorithm design processes is limited, particularly for scalarization methods. Even a slight alteration in the objective function's form or problem scale is highly likely to result in failure for the weights or weight generators specifically designed for the current MOO problem.
For instance, in the case of problems with similar scales, the average weighted gradient produces updates that tend to align in a consistent descent direction. In the presence of disparate gradient scales, this gradient may be disproportionately influenced by the largest gradient magnitude while disregarding other objectives.
These difficulties are actually attributed to the conflict in objective functions, which makes it hard to design update direction generators. However, some manually designed MOO methods based on weighted combinations have achieved excellent results in some problems, due to the researchers' profound understanding of the problems.

\begin{figure}[h!]
\centering
\includegraphics[scale=0.4]{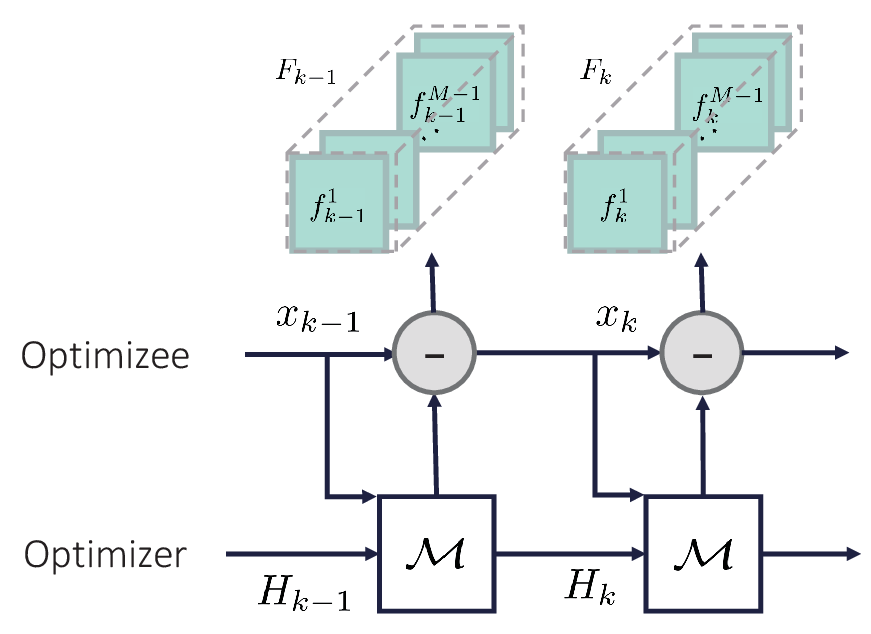}
\caption{Diagram of ML2O.  $\mathcal{M}$ denotes the learning optimizer, ${F}$ represents the objective function of MOO and $H$ signifies the historical information of the iteration trajectory.}
\label{fig:intro}
\end{figure}

With the rapid advancement of machine learning, artificial intelligence (AI) has surpassed expectations in comprehending informations \cite{gpt_4, vaswani2017attention,dosovitskiy2020image, shu2023learning}.
Due to the fact that the design process of optimization methods can be seen as researchers' understanding of the changes in objective function values, using AI's powerful understanding ability to replace human cognition and generate an optimization method is very attractive.
Designing optimization methods by AI necessitates the transformation of the human-designed process into a learnable problem, and to learn a generator (or mapping) that can effectively determine the update direction based on the gradient of the objective function.
This transformation as a new paradigm has been proposed in the single-objective optimization and has been demonstrated successful applications across various machine learning optimization tasks \cite{wang2019hyperadam,heaton2020safeguarded,lv2017learning,chen2017learning}. This approach is a mathematical embodiment of the ideas of AI for science, and the single-objective case is known as learning to optimize (L2O) \cite{andrychowicz2016learning}.
Inspired by L2O, we can treat each step of the iterative MOO method as a layer of a neural network, and the process of optimizing the MOO problem can be considered as a learnable recurrent neural network that learns a method for generating update directions from meta-data. In this paper, we propose a learnable generator of MOO update directions, called the multi-gradient learning to optimize (ML2O) method.
The iterative process of ML2O is illustrated in Figure \ref{fig:intro}.
This AI-driven process of conceptualizing and generating optimizers can be regarded as a new paradigm for designing MOO methods. Our proposed ML2O method has two advantages:

${ \bullet }$ ML2O replaces human intervention to comprehend the distinctions and interconnections among multiple objective functions by machine learning, relieving researchers from the arduous task of designing a MOO method.

${ \bullet }$ With the capability of neural networks to approximate arbitrary functions, ML2O is no longer constrained to weighted combinations of gradients from conventional methods when learning a generator in the update direction. This presents an opportunity for exploring a larger function space and achieving more intricate mappings.

Morevoer, we provide the convergence analysis of ML2O and propose a guarded multiple gradients learning to optimize (GML2O) method by incorporating the ideal of safeguard \cite{premont2022simple,heaton2020safeguarded}. Considering that safeguard necessitates a convergent method, we develop a stochastic MOO method called dynamic sampling stochastic multiple gradients (DSSMG) method and establish its convergence to a Pareto critical point. Based on DSSMG, we demonstrate that the iterative sequence generated by the GML2O method converges towards a Pareto critical point.
Experimental results on MTL demonstrate that our ML2O possesses the capability to learn an effective direction for minimizing the loss of all tasks, surpassing conventional scalarization and gradient-based methods in terms of final performance. Our generalizability experiments conducted on various optimization step settings, network structure configurations and datasets, further validate the superiority of our approach over manually designed methods by showcasing its exceptional generalization ability. These findings collectively underscore the outstanding performance of ML2O in optimizing MOO problems. This show that our proposed paradigm has potential to provide a promising avenue for future research.

This paper is organized as follows. Section \ref{sec:preliminaries} presents some basic definitions,
notations and some MOO methods. Section \ref{sec:ML2O} formally introduce the structure and methodology of ML2O. We propose GML2O with a guarded criterion and prove convergence of GML2O in Section \ref{sec:gml2o}.
Experimental results of ML2O and GML2O training MTL neural networks are shown in Section \ref{sec:experment}.
Finally, some conclusions are included in the last section of the paper

\section{Preliminaries}
\label{sec:preliminaries}
Let us first present some notations that will be used in this paper. Let ${{\mathbb{R}^N}}$ denote the $N$-dimensional real Euclidean space, $M$ be the positive integer greater than 1 and $-\mathbb{R}_{++}^M:=\left\{-u: u \in \mathbb{R}_{++}^M\right\}$. Consider the following MOO problem
\begin{align}\label{eq:moo}
\min _{{x} \in { {\mathbb{R}^N}} } {{F}}({x})=\left(f^1({x}), \ldots, f^M({x})\right)^{T},
\end{align}
where ${F:{\mathbb{R}^N} \to {\mathbb{R}^M}}$ is a vector-valued function, and each $f^i({x})$, ${i \in \left[ M \right] = \left\{ {1, \cdots ,M} \right\}}$ is continuously and differentiable.
We denote by
\[\nabla F = {\left( {\nabla {f^1}(x), \cdots ,\nabla {f^M}(x)} \right)^T}, \]
the Jacobian matrix associated with $F$.
The concept of optimality for multi-objective optimization problems (\ref{eq:moo}) is now introduced. In contrast to single-objective optimization, which aims to identify the minimum or maximum point of an objective function, MOO involves investigating tradeoffs and interdependencies among multiple objectives.
This shift allows for a broader exploration of the solution space, considering the interplay between different objectives and the potential for achieving a set of solutions that are not dominated by each other.
Such points that cannot be further improved are referred to as Pareto optimality.
\begin{definition}(Pareto optimality)
A point $x^* \in \mathbb{R}^N$ is a Pareto optimal solution for problem (\ref{eq:moo}) if there does not exist $x \in \mathbb{R}^N$ such that $f^i(x) \leq f^i\left(x^*\right)$ for all $i \in \left[ M \right]$ , and $f^{i_0}(x)<f^{i_0}\left(x^*\right)$ for at least one $i_0 \in \mathbb{R}^N$.
\end{definition}

Optimizing MOO is an attempt to locate a Pareto optimality \cite{custodio2011direct}, and a necessary condition for it is that a point $x^* \in \mathbb{R}^N$ is Pareto criticality:
$$
\nabla F\left( {x^*} \right)\left(\mathbb{R}^N\right) \cap\left[-\mathbb{R}_{++}^M\right]=\emptyset,
$$
where $\nabla F\left( x \right)\left(\mathbb{R}^N\right):=\left\{\nabla F\left( \bar{x} \right) v: v \in \mathbb{R}^N\right\}$.
Thus we have the following definition.
\begin{definition}\label{def:Pareto critical}(Pareto criticality)
${x^*} \in \mathbb{R}^N$ is Pareto critical for problem (\ref{eq:moo}) if there is no direction ${d \in {\mathbb{R}^N}}$ such that ${\nabla {f^i}{(x^*)^T}d < 0}$ for all ${i \in \left[ M \right]}$.
\end{definition}
It is evident from this definition that if $x^*$ is not a Pareto critical point, then there exists a direction $d$ such that every objective of $F$ exhibits local decreasing behavior at the point $x^*$. Therefore, exploring $d$ in a local neighborhood of $x^*$ through iteration leads to an improved solution dominating $x^*$ \cite{zhou2022convergence}.
As Pareto criticality is indicative of a local property, it is commonly utilized as metrics for investigating local minima in MOO problems with non-convex objective functions \cite{fliege2019complexity}.

The key challenge posed by the MOO problem lies in the conflicting objectives, i.e. ${\nabla {f^i}{(x)^T}\nabla {f^j}(x) < 0}$, ${i \ne j}$, ${i,j \in \left[ M \right]}$.
To effectively address conflicts in objectives, the most commonly-used approach in research are linear scalarization and MOGM, which iteratively yield sequences ${\left\{ {{x_k}} \right\}}$ with the following procedure:
\begin{align*}
{x_{k + 1}} = {x_k} + {\alpha _k}{d_k},
\end{align*}
where ${{d_k}}$ is a search direction, and ${{\alpha _k}}$ is a step size.
Linear scalarization \cite{lin2019pareto, liu2021conflict, liu2019end} uses a linear weighted sum method to combine the function value of all objectives:
\[\bar F\left( {{x_k}} \right) = \sum\limits_{i = 1}^M {{\lambda ^i}{f^i}({x_k})} \]
where ${{{\lambda ^i}}}$ is the weight for the $i$-th objective.
This type of method is straightforward and extensively employed, however, it necessitates the manual allocation of weights lambda in the optimizer, which poses significant challenges.
As for MOGM, by calculating the first-order Taylor expansion approximation at point ${{{x_{k + 1}}}}$ for each target ${i \in \left[ M \right]}$, we can obtain a measure of the descent for each objective as
\[{f^i}({x_k}) - {f^i}({x_k} + {\alpha _k}{d_k}) \approx  - {\alpha _k}d_k^T\nabla {f^i}({x_k}).\]
Therefore, if a direction can be identified that satisfies ${ - d_k^T\nabla {f^i}({x_k}) > 0}$ for all objectives, it implies that this particular direction leads to the descent of all objective functions. Numerous approaches exist to obtain such directions, in the next we elucidate the conventional multiple-gradient descent algorithm (MGDA) \cite{fliege2000steepest}.
By utilizing Definition \ref{def:Pareto critical}, MGDA can be directly optimized towards the Pareto critical point \cite{zhou2022convergence}. Specifically, at every ${k \in \left[ K \right]}$ step, we denote the update direction ${d_k}$ generated by MGDA as ${ d\left( {{x_k}} \right) }$
which can be derived by solving the subsequent subproblem
\begin{align}\label{eq:mgda_1}
d\left( x_k \right) = \mathop {\arg \min }\limits_{d \in {\mathbb{R}^N}} \mathop {\max }\limits_{i \in [M]} \left\{ {\nabla {f^i}{{(x_k)}^T}d} \right\} + \frac{1}{2}{\left\| d \right\|^2},
\end{align}
where ${{\alpha _k}{\nabla {f^i}{{(x_k)}^T}d}}$ is the first-order Taylor approximation of ${\nabla {f^i}\left( {{x_k}} \right) - \nabla {f^i}\left( {{x_k} + {\alpha _k}d\left( {{x_k}} \right)} \right)}$. Observed that the objective function is proper, closed and strongly convex, this problem has always a unique optimal solution.  In order to obtain this solution, we consider the dual form of this subproblem as a min-norm oracle
\begin{align}
{\lambda ^*}\left( x \right) &\in \mathop {\arg \min }\limits_{\lambda  \in {\Delta _M}} {\left\| {\nabla F(x)\lambda } \right\|^2},\label{eq:mgda_2_2}
\end{align}
where ${{\Delta ^M} = \left\{ {\lambda |\sum\limits_{i = 1}^M {{\lambda ^i} = 1,{\lambda ^i} \ge 0~~\forall i} } \right\}}$ denotes the simplex set. The direction ${d\left( {{x_k}} \right)}$ is given by
\begin{align}
d\left( x_k \right) &= \sum\limits_{i = 1}^M {{\lambda ^{*i}}\left( x_k \right)\nabla {f^i}\left( x_k \right)} \label{eq:mgda_2_1}.
\end{align}
Such full-gradient manipulation algorithms to solve MOO have theoretical guarantees \cite{yu2020gradient, fukuda2014survey}, but imposes a tremendous computational consumption.
In practical applications, the stochastic gradient method is used more widely. Without loss of generality, we consider a MOO problem over an input space ${\Xi}$, we obtain noisy gradient feedback
\[\nabla F\left( {{x_k},{\xi _k}} \right): = \left( {\nabla {f^1}({x_k},{\xi _k}), \cdots ,\nabla {f^M}({x_k},{\xi _k})} \right)^T,\]
where stochastic noise ${{\xi _k} \in \Xi }$ is i.i.d sampled, $M$ is the number of objectives.
In machine learning especially supervised learning, ${\xi_k}$ usually represents a batch of labeled data samples taken from the whole data set, for which there is no explicit form since the feature and label pairs are drawn according to an unknown distribution.

Replaces the full gradient with the stochastic version ${\nabla {f^i}(x_k,\xi )}$ in (\ref{eq:mgda_1}),
Liu et al. \cite{liu2021stochastic} proposed stochastic counterpart of MGDA, referred to as stochastic multiple gradient (SMG) manipulation.
Update direction ${d_k}$ of the ${k}$th step with noise ${{{\xi _k}}}$ is denoted as ${d\left( {{x_k},{\xi _k}} \right)}$. We have that
\[d\left( {{x_k},{\xi _k}} \right) = \mathop {\arg \min }\limits_{d \in {\mathbb{R}^N}} \mathop {\max }\limits_{i \in [M]} \left\{ {\nabla {f^i}{{({x_k},{\xi _k})}^T}d} \right\} + \frac{1}{2}{\left\| d \right\|^2},\]
and the iteration is as following:
\[{x_{k + 1}} = {x_k} - {\alpha _k}d\left( {{x_k},{\xi _k}} \right),\]
where ${{\alpha _k}}$ is the step size of the $k$th step.
At this point, the subproblem of solving for the descent direction becomes
\begin{align}
{\lambda ^*}\left( {x_k,\xi_k } \right) &\in \mathop {\arg \min }\limits_{\lambda  \in {\Delta _M}} {\left\| {\nabla F(x_k,\xi_k )\lambda } \right\|^2}, \label{eq:smg_lamda}
\end{align}
then the direction ${d\left( {x_k,\xi_k } \right)}$ is calculated by
\begin{align}
d\left( {x_k,\xi_k } \right) &= \sum\limits_{i = 1}^M {{\lambda ^{*i}}\left( {x_k,\xi_k } \right)\nabla {f^i}\left( {x_k,\xi_k } \right)}. \label{eq:smg_dk}
\end{align}
SMG is a straightforward method used to find the update direction in the stochastic case, the introduction of noise raise issues in the biasedness in the stochastic multi-gradient manipulation.
Unlike the single objective optimization introduced assumption of unbiased gradients guarantees the effectiveness of the expected descent, stochastic gradients in MOO may lead to the failure of multi-gradient operations.
Fernnado et al. \cite{fernando2023mitigating} have pointed out that the solution of the subproblem (\ref{eq:smg_lamda}) is nonlinear in ${\nabla F\left( {{x_k},{\xi _k}} \right)}$.
Thus despite the assumption that all gradients are unbiased ${{\mathbb{E}_{{\xi _k}}}\left[ {\nabla {f^i}({x_k},{\xi _k})} \right] = \nabla {f^i}({x_k})}$, we still have the fact that ${{\mathbb{E}_{{\xi _k}}}\left[ {d\left( {{x_k},{\xi _k}} \right)} \right] \ne d\left( {{x_k}} \right)}$.

To overcome the limitations of gradient algorithms in handling noise estimation, stochastic gradient algorithms for single-objective optimization have explored two main approaches for noise reduction \cite{bottou2018optimization}. The first approach is gradient aggregation methods, which leverage historical information stored during iterations to enhance the quality of the search direction and improve the current iteration. The second approach is dynamic sampling methods \cite{yong}, which progressively increase the number of stochastic gradient samples and utilize increasingly accurate gradients in the optimization process to reduce noise. These methods have also been applied to MOO.

Fernando et al. \cite{fernando2023mitigating} proposed a momentum-like gradient manipulation to addresses the interference caused by stochastic gradients in algorithms which updated as follows, for all ${i \in \left[ M \right]}$
\[y_{k + 1}^i = {\Pi _{{{\cal Y}_i}}}\left( {{\beta _k}\nabla {f^i}({x_k},{\xi _k}) + \left( {1 - {\beta _k}} \right)y_k^i} \right),\]
\[{\lambda _{k + 1}} = {\Pi _{{\Delta _M}}}\left( {{\lambda _k} - {\gamma _k}\left( {Y_k^T{Y_k} + \rho I} \right){\lambda _k}} \right), \]
\[d_k = \sum\limits_{i = 1}^M {\lambda _k^i{y_k}}, \]
where ${{\beta _k}}$ and ${{{\gamma _k}}}$ are both hyperparameters, ${y_{k + 1}^i}$ is the ``tracking" variable obtained by ${{\nabla {f^i}({x_k},{\xi _k})}}$, ${I \in {\mathbb{R}^{M \times M}}}$ is the identity matrix, ${{\Pi _{{{\cal Y}_i}}}}$ and ${{\Pi _{{\Delta _M}}}}$ are denotes the projection to a bounded set ${{\cal Y}_i}$ and a probability simplex ${{{\Delta _M}}}$ respectively.
By exponentially averages the past calculated weights, Zhou et al. \cite{zhou2022convergence} proposed a composite weights determination scheme to integrate historical information into the weights.
The modified weighting formula is as follows
\[{\lambda _k} = {\beta _k}{\lambda _{k - 1}} + \left( {1 - {\beta _k}} \right){\lambda ^*}\left( {{x_k},{\xi _k}} \right),\]
\[d_k = \sum\limits_{i = 1}^M {\lambda _k^i\nabla {f^i}\left( {{x_k},{\xi _k}} \right)}. \]
As the number of iterations increases and the solution approaches the Pareto optimal solution, ${{\beta _k}}$ tends to increase, while maintaining a relatively stable ${{\lambda _k}}$ helps eliminate strong correlation and ensures convergence and performance.

In fact, all of the aforementioned methods for determining the update direction $d_k$ can be simplified to this mapping from multiple gradients to update directions by
\begin{align}\label{eq:humen}
{d_k} = \mathcal{M}\left( {\left\{ {\nabla {f^i}({x_k},{\xi _k})} \right\}_{i = 1}^M,{H_k}} \right),
\end{align}
where ${H_k}$ represents the historical information of the iteration trajectory, and ${\mathcal{M}}$ denotes a mapping of update directions, implying human involvement in subproblem design, solving subproblem, and hyperparameter selection.

\section{ML2O}\label{sec:ML2O}

\begin{figure}[h!]
\centering
\includegraphics[scale=0.45]{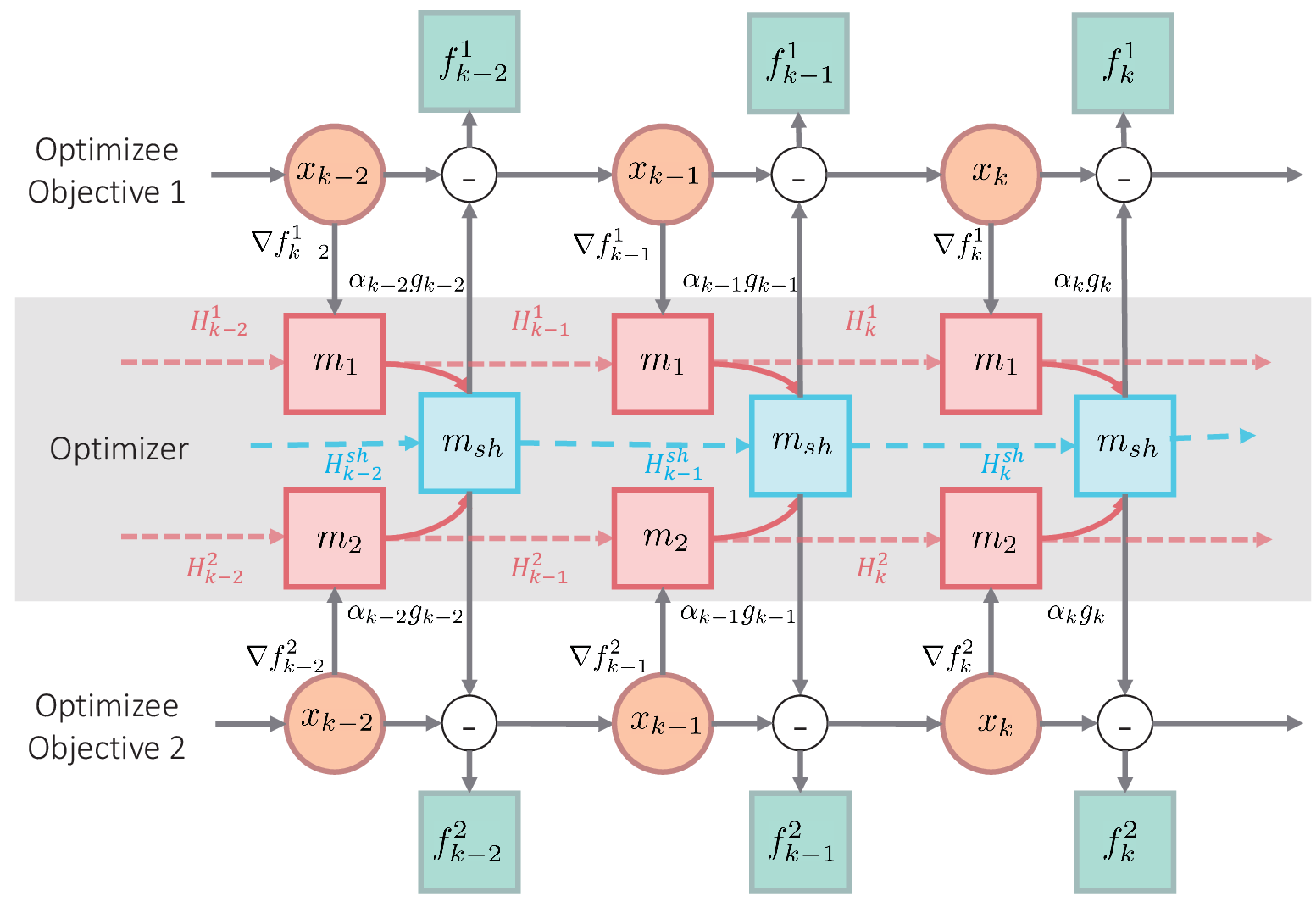}
\caption{Computational graph of ML2O at two objectives. The ML2O framework consists of specific-modules ${{m_1}}$ and ${{m_2}}$ along with the shared-module ${{m_{sh}}}$. ${H}_k^1$, ${H}_k^2$ are the history information of specific-modules, ${{H}_k^{sh}}$ is the history information of shared modules. In each training step $k$, ${\alpha_k}$ is the step size.}
\label{fig:ml2o}
\end{figure}

The acquisition of update directions for multiple objectives necessitates a substantial time investment on the part of researchers, owing to the imperative consideration of interactions and disparities among these objectives.
We propose ML2O, a learning-based optimizer that effectively learns a mapping from multi-gradient and historical information to determine update directions $g  _k$, which can be expressed as follows
\begin{align}\label{eq:ai}
{{g_k} = \mathcal{M}\left( {\left\{ {\nabla {f^i}({x_k},{\xi _k})} \right\}_{i = 1}^M,{H_k};\Theta } \right)}.
\end{align}
In contrast to (\ref{eq:humen}), the learning optimizer incorporates a parameter $\Theta$ into the gradient mapping design, where different values of $\Theta$ represent various optimizers.
The objective function of MOO as a learning machine is refer to as \textit{learner}, the objective function value of MOO is defined as \textit{optimizee}, and the optimization algorithm that minimizes optimizee is called \textit{optimizer}.
ML2O is designed as a learning-based optimizer with two main components, one shared-module ${m_{sh}}$ and $M$ specific-modules ${m_{i}}$, ${i \in \left[ M \right]}$, where ${m_{sh}}$ is a generic neural architecture shared between different objectives, and ${{m_i}}$ are objective specific neural architectures, parallel and independent of each specific objective.
Figure \ref{fig:ml2o} gives an overview of ML2O framework in the case of two objectives.

%

For the $k$th iteration and the $i$th objective, we input the gradient ${\nabla {f_k^i}}$ and state metric ${H_k^i}$ into the specific-module $m_i$ to obtain the feature ${{h_k^i}}$, and combine the ${{\left\{ {h_k^i} \right\}_{i \in \left[ M \right]}}}$ into a feature matrix ${s_k^{sh}}$. Then the state metric ${H_k^{sh}}$ of the shared module together with ${s_k^{sh}}$ are input into the shared module ${m_{sh}}$ to get the update direction ${g_k}$ of ${x_k}$.
The iterative trajectory information integrates data from historical gradients that acts similarly to momentum.
It is well known that historical information has been demonstrated to be valid in the optimization of neural networks (e.g., adaptive moment estimation (Adam) method \cite{kingma2014adam}).
In particular, the utilization of historical information in stochastic gradient updates commonly employed in MOO can effectively alleviate the bias of gradients \cite{fernando2023mitigating, zhou2022convergence}.
To enable the learning optimizer to incorporate historical information and process sequential data, we employ a recurrent neural network with long short-term memory LSTM as the structure for ${m_i}$ and ${m_{sh}}$.
It is worth noting that we only detail the case where each module is a single-layer LSTM, the multi-layer direct case can be obtained similarly by directly stacking single-layer LSTM.
In the sequel, we will describe in detail the composition as well as the function of each module in ML2O.

\textbf{Specific-module} ~~Module ${m_i}$, ${i \in \left[ M \right]}$ as objective specificed are designed to handle each individual objective in parallel and feed the output into the shared-module.
The LSTM units in ${m_i}$ with hidden size $H$ at each $k$ step consisting an \textit{input gate} ${i_k^{{m_i}}}$, a \textit{forget gate} ${f_k^{{m_i}}}$, an \textit{output gate} ${o_k^{{m_i}}}$, a \textit{memory cell} ${c_k^{{i}}}$ and a \textit{hidden state} ${h_k^{{i}}}$. ${\sigma}$ is the sigmoid function, and ${\odot }$ is the Hadamard product. The specific-module ${m_i}$ is compactly specified as follows,
\begin{align*}
i_k^{{m_i}} &= \sigma \left( {W_p^is _k^i + b_p^i + W_{hp}^ih_{k - 1}^i + b_{hp}^i} \right),\\
f_k^{{m_i}} &= \sigma \left( {W_f^is _k^i + b_f^i + W_{hf}^ih_{k - 1}^i + b_{hf}^i} \right),\\
g_k^{{m_i}} &= \tanh \left( {W_f^is _k^i + b_f^i + W_{hf}^ih_{k - 1}^i + b_{hf}^i} \right),\\
o_k^{{m_i}} &= \sigma \left( {W_o^is _k^i + b_o^i + W_{ho}^ih_{k - 1}^i + b_{ho}^i} \right),\\
c_k^{{i}} &= f_k^{{m_i}} \odot c_{k - 1}^{{i}} + i_k^{{m_i}} \odot g_k^{{m_i}},\\
h_k^{{i}} &= o_k^{{m_i}} \odot \tanh \left( {c_k^{{i}}} \right),
\end{align*}
where ${s_k^i }$ is generated by gradient ${{\nabla {f^i}}}$ after preprocessing. The update of each LSTM unit can be written precisely as follows,
\[\left[ {\begin{array}{*{20}{c}}
{h_k^i}\\
{c_k^i}
\end{array}} \right] = {m^i}\left( {s_k^i,H_k^i;{\Theta ^i}} \right),\]
where ${H_k^i = \left\{ {h_k^i,c_k^i} \right\}}$ and ${{\Theta ^i} = \left\{ {{W^i},{b^i}} \right\}}$ , ${{W^i}}$ and ${b^i}$ are the parameters of task-specific module ${m^i}$.

\textbf{Shared-module } Shared-module ${m_{sh}}$ is designed to exploit the shared information between these different objectives, which takes the output ${s_k^{sh} = \left( {h_k^{{1}}, \cdots ,h_k^{{M}}} \right)}$ of the specific-module as input. The shared-module uses the same LSTM architecture as specific-module, such that the formulaic expression for ${m_{sh}}$ is
\begin{align*}
\left[ {\begin{array}{*{20}{c}}
{h_k^{sh}}\\
{c_k^{sh}}
\end{array}} \right] = {m^{sh}}\left( {s_k^{sh},H_k^{sh};{\Theta ^{sh}}} \right),
\end{align*}
where ${H_k^{sh} = \left\{ {h_k^{sh},c_k^{sh}} \right\}}$, ${h_k^{sh}}$ and ${c_k^{sh}}$ are hidden states and memory cell of the shared-module, respectively. Considering the fact that shared-module need to process information from $M$ objectives, we set the hidden size of the LSTM unit of shared-module is $MH$ to be $M$ times specific-module.

Having obtained the output of the shared-module ${{h_k^{sh}}}$, we adopt a linear layer to map ${h_k^{sh}}$ to ${g_k}$, completing the final step of generating the update direction like
\[{g_k} = liner\left( {h_k^{sh};{\Theta ^l}} \right),\]
the output ${g_k}$ of the linear layer is the updated direction of ${x_k}$ obtained by ML2O.

In our ML2O, the specific-module ${m_i}$ can be regarded as pre-processing module and mapping gradients of different objectives to a high-dimensional feature space. Extracting information that facilitates the optimization of the current objective irrespective of other objectives. The shared-module ${m_{sh}}$ then synthesizes information from different objectives in a high-dimensional feature space to find the update direction for the current iteration step.

Summarizing the above discussion, we designed ML2O with adequate consideration of the structural characteristics of multi-objective and combined L2O with MOO optimization.
In the next, we investigate approaches for training the learning optimizer $\mathcal{M}$ in order to effectively discover a high-quality solutions to MOO.

\textbf{learning ML2O} ~~In order to optimize the parameters of ML2O neural network, we introduce the following loss function
\[{L_k}\left( {{\Theta}} \right) = \mathop {\max }\limits_{i \in \left[ M \right]} \left\{ {{f^i}\left( {{x_k},{\Theta}} \right) - {f^i}\left( {{x_{k - 1}},{\Theta}} \right)} \right\},\]
where ${x_k}$ is the parameter of learner at $k$ step and ${\Theta}$ is the parameter of ML2O.
Minimizing this maximum value is a crucial step in optimization, aimed at discovering the optimal common descent direction \cite{fukuda2014survey}. We use it to guide the update of the ML2O method.
Furthermore, we adopt back propagation through time \cite{werbos1990backpropagation} to optimize ML2O parameters ${\Theta  = \left\{ {{\Theta ^1}, \cdots ,{\Theta ^M},{\Theta ^{sh}},{\Theta ^l}} \right\}}$.
Specifically, when updating the parameter $\Theta$, $K$ steps are split into $T$ periods of ${\bar K}$ steps with ${T = \frac{{K}}{\bar K}}$, each period ${t \in \left[ T \right] = \left\{ {1, \cdots ,T} \right\}}$, parameter ${{{\Theta _t}}}$ optimized by minimizing the averaged regret on the meta-train set with step size ${\alpha_k}$:
\begin{align*}
&\mathcal{L}\left( {{\Theta _t}} \right) = \frac{1}{{\bar K}}\sum\limits_{k = 1}^{\bar K} {{L_k}\left( {{\Theta _t}} \right)}, \\
&{\Theta _{t + 1}} = {\Theta _t} - {\alpha _k}\frac{{\partial \mathcal{L}\left( {{\Theta _t}} \right)}}{{\partial {\Theta _t}}}.
\end{align*}
This implies that for each update of the learning optimizer $\mathcal{M}$ with respect to parameter ${\Theta _t}$, $x_k$ is first updated in $\bar K$ steps.
It is worth noting that, the ML2O method outperforms designed optimizers in terms of performance, but it lacks the ability to prove convergence. In order to further investigate the convergence of ML2O, in the next section we propose the ML2O method with a guarding mechanism.
\section{GML2O} \label{sec:gml2o}
The guarding mechanism is a methodology that aims to investigate convergence by introducing specific design criterion. It operates by taking updates obtained from the learning optimizer ML2O, as well as algorithms known to exhibit convergence, and selectively filtering these updates based on predefined guarding criterion. When the learning optimizer performs well, it is utilized; however, if its performance deteriorates, the design optimizer is activated.
The dynamic switching between optimizers ensures that the resulting choice of update direction is at least as good as that of the designed algorithm, while simultaneously guaranteeing convergence of the learning to optimize method.

We introduce the dynamic sampling stochastic multiple gradient (DSSMG) method and formal proof of its convergence, and provide a comprehensive description of DSSMG and provide a formal proof of its convergence, thereby demonstrating its capacity to reach the Pareto criticality. Given its extensive practical utilization, this method holds significant importance as a safety criterion for designing algorithms that ensure the convergence of GML2O.

\subsection{DSSMG} \label{sec:smg}

We proposed DSSMG method, a approach requires dynamically selecting a predetermined number of stochastic gradients during each iteration, where the number of sampled gradients increases proportionally with the progress of iteration steps.
Then, the average value of all sampled gradients is computed, serving as an approximation to the complete gradient.

Without loss of generality, for all ${q > 0}$ and ${N_B > 0}$, assume that the dynamic sample size of DSSMG method is defined as
\begin{align}\label{eq:sample_size}
{N_k} = \max \left\{ {{N_B},{k^q}} \right\},
\end{align}
where $N_B$ is the threshold for dynamic sample size, and the rate of sample increase is governed by the parameter $q$. It is desirable for $q$ to be sufficiently small, yet non-zero, in order to avoid an excessively rapid growth of the sample size in dynamic sampling.
Assume that the DSSMG as ${{Y_k} = \left( {y_k^1, \cdots ,y_k^M }\right)^T }$, for all ${i \in \left[ M \right]}$ we have
\begin{align}\label{eq:yk}
{y_k^i} = \frac{1}{{{N_k}}}\sum\limits_{j = 1}^{{N_k}} {\nabla {f^i}\left( {{x_k},{\xi _{k,j}}} \right)}.
\end{align}
Following (\ref{eq:mgda_1}), we have
\begin{align} \label{eq:mini_batch_dk}
{d_k} = \mathop {\arg \min }\limits_{d \in {\mathbb{R}^N}} \mathop {\max }\limits_{i \in [M]} \left\{ {{\left( {y_k^i} \right)^T}d} \right\} + \frac{1}{2}{\left\| d \right\|^2}.
\end{align}
As with the discussion of (\ref{eq:mgda_2_2}) and (\ref{eq:mgda_2_1}), we have
\begin{align}
\lambda _k^i &\in \mathop {\arg \min }\limits_{\lambda  \in {\Delta _M}} {\left\| {\sum\limits_{i = 1}^M {{\lambda ^i}{y_k}} } \right\|^2}, \label{eq:msmoo_lamda}
\end{align}
and
\begin{align}
{d_k} &= \sum\limits_{i = 1}^M {\lambda _k^iy_k^i}, \label{eq:msmoo_dk}
\end{align}
where $d_k$ is the update direction of the $k$th iteration, and (\ref{eq:msmoo_lamda}) is solved by the Frank-wolf method \cite{sener2018multi}. We summarize the DSSMG method in Algorithm~\ref{alg:mini_batch_SMG}.

\begin{algorithm}[h]
\begin{normalsize}
\caption{\text {DSSMG method}}
\begin{algorithmic}\label{alg:mini_batch_SMG}
\REQUIRE Objective function of MOO ${F = \left( {{f_1}, \cdots ,{f_M}} \right)^T}$, number of iterations $K$, step size $\alpha_k$, dynamic sampling parameter $q$ and threshold $N_B$.
\ENSURE ${x_0}$
\FOR{${k = 1, \cdots ,K}$}
\STATE ${{N_k} = \max \left\{ {{N_B},{k^q}} \right\}}$
\STATE ${{y_k^i} = \frac{1}{{{N_k}}}\sum\limits_{j = 1}^{{N_k}} {\nabla {f^i}\left( {{x_k},{\xi _{k,j}}} \right)}}$
\STATE ${{d_k} = \mathop {\arg \min }\limits_{d \in {\mathbb{R}^N}} \mathop {\max }\limits_{i \in [M]} \left\{ {{\left( {y_k^i} \right)^T}d} \right\} + \frac{1}{2}{\left\| d \right\|^2}}$ \COMMENT{Solved by (\ref{eq:msmoo_lamda}) and (\ref{eq:msmoo_dk}) }
\STATE ${{x_{k + 1}} = {x_k} - {\alpha _k}d_k}$
\ENDFOR
\end{algorithmic}
\end{normalsize}
\end{algorithm}

\begin{remark}
It is worth noting that we have the flexibility to choose $q$ as any number greater than zero. Therefore, the DSSMG method can make ${{N_B} \ge {k^q}}$ constant by choosing the appropriate $q$ within finite steps, at which point DSSMG becomes a mini-batch gradient method with batch size ${N_B}$. The number of iterations $K$ is finite in most practical applications, so it is of practical significance for us to propose $N_k$.
\end{remark}
Then, we provide an analysis of the convergence of DSSMG method.
First, we make the following assumptions.
\begin{assumption}\label{ass_1}
For all objective functions ${{{f^i}\left( x_k \right)}}$, ${i \in [M]}$, iterates ${k \in [K]}$ and i.i.d stochastic variable ${\xi  \in \Xi }$, we have access to the individual stochastic gradients ${{\nabla {f^i}\left( {{x_k},{\xi}} \right)}}$ which is unbiased estimates of ${{\nabla {f^i}\left( {{x_k}} \right)}}$, i.e. ${\nabla {f^i}\left( {{x_k},\xi } \right) = {\mathbb{E}_{{\xi}}}\left[ {\nabla {f^i}\left( {{x_k},\xi } \right)} \right]}$ and each gradient variance is bounded by ${{\sigma ^2}}$, i.e.,
\[{\mathbb{E}_{\xi} }\left[ {{{\left\| {\nabla {f^i}({x_k},\xi ) - \nabla {f^i}\left( {{x_k}} \right)} \right\|}^2}} \right] \le {\sigma ^2}.\]
\end{assumption}

\begin{assumption}\label{ass_2}
For all objective functions ${{{f^i}\left( x_k \right)}}$, ${i \in [M]}$ and iterates ${k \in [K]}$, the following hold

(a) ${{{f^i}({x_k})}}$ is bounded from below
\[\mathbb{E}\left[ {{f^i}({x_k})} \right] \ge {F_{\inf }} >  - \infty. \]

(b) ${f^i(x_k)}$ is differentiable at every point $x_k$, and the gradient ${{\nabla {f^i}({x_k})}}$ is bounded.

(c) ${{\nabla {f^i}}}$ is Lipschitz continuous with constant $L$.
\end{assumption}
\begin{remark}
According to Assumption \ref{ass_2}(b) it can be easily obtained that there exists a positive constant $C_1 \ge 0$ such that ${\left\| {\nabla {F_k}} \right\| \le C_1}$. Therefore, the Jacobian matrix of $F$ at ${x_k}$ is bounded.
\end{remark}
It is worth noting that, Liu et al. \cite{liu2021stochastic} assumed that the optimal solution ${{\lambda ^*}\left( {x_k,\xi } \right)}$ of problem (\ref{eq:smg_lamda}) is Lipschitz continuous with gradient ${{\nabla F(x_k,\xi )}}$, which has been demonstrated to be unreasonable \cite{zhou2022convergence}. We next provide convergence results for DSSMG method under Assumptions~\ref{ass_1} and \ref{ass_2} without dependent on this Lipschitz continuous assumption. Then, we establish the following lemmas.

\begin{lemma}\label{lamma_1}
Suppose that Assumption \ref{ass_1} holds, then for each ${i \in [M]}$ and ${\xi_{k,j}  \in \Xi }$, the sequence of iterates ${{\left\{ {y_k^i} \right\}_{k \in \left[ K \right]}}}$ satisfies
\[{\mathbb{E}_k}\left[ {y_k^i} \right] = \nabla {f^i}\left( {{x_k}} \right),{\mathbb{E}_k}\left[ {{{\left\| {y_k^i - {\mathbb{E}_k}\left[ {y_k^i} \right]} \right\|}^2}} \right] \le \frac{1}{{{N_k}}}{\sigma ^2},\]
where ${{\mathbb{E}_k}\left[  \cdot  \right]}$ represents the expectation for all the stochastic variables ${\xi_{k,j}}$ in the $k$th step, ${{y_k^i}}$ is the dynamic sampling stochastic multiple gradient defined by (\ref{eq:yk}).
\end{lemma}
\emph{proof:}
From Assumption \ref{ass_1} and the definition of ${{y_k^i}}$, it is straightforward to obtain
\begin{align*}
{\mathbb{E}_k}\left[ {y_k^i} \right] &= {\mathbb{E}_k}\left[ {\frac{1}{{{N_k}}}\sum\limits_{j = 1}^{{N_k}} {\nabla {f^i}\left( {{x_k},\xi _{k,j}} \right)} } \right]\\
& = \frac{1}{{{N_k}}}\sum\limits_{j = 1}^{{N_k}} {{\mathbb{E}_k}\left[ {\nabla {f^i}\left( {{x_k},\xi _{k,j}} \right)} \right]} \\
& = \frac{1}{{{N_k}}}\sum\limits_{j = 1}^{{N_k}} {\nabla {f^i}\left( {{x_k}} \right)}  = \nabla {f^i}\left( {{x_k}} \right).
\end{align*}
Similarly, we have
\begin{align*}
{\mathbb{E}_k}\left[ {{{\left\| {y_k^i - {\mathbb{E}_k}\left[ {y_k^i} \right]} \right\|}^2}} \right] &= {\mathbb{E}_k}\left[ {{{\left\| {y_k^i - \nabla {f^i}\left( {{x_k}} \right)} \right\|}^2}} \right]\\
& = {\mathbb{E}_k}\left[ {{{\left\| {\frac{1}{{{N_k}}}\sum\limits_{j = 1}^{{N_k}} {\nabla {f^i}\left( {{x_k},{\xi _{k,j}}} \right)}  - \nabla {f^i}\left( {{x_k}} \right)} \right\|}^2}} \right]\\
& = \frac{1}{{N_k^2}}\sum\limits_{j = 1}^{{N_k}} {{\mathbb{E}_k}\left[ {{{\left\| {\nabla {f^i}\left( {{x_k},{\xi _{k,j}}} \right) - \nabla {f^i}\left( {{x_k}} \right)} \right\|}^2}} \right]}  \\
&~~~~+ \frac{1}{{N_k^2}}\sum\limits_{1 \le l < u \le {N_k}}^{} {Cor\left( {\nabla {f^i}\left( {{x_k},{\xi _{k,l}}} \right),\nabla {f^i}\left( {{x_k},{\xi _{k,u}}} \right)} \right)} \\
& \le \frac{1}{{N_k^2}}\sum\limits_{j = 1}^{{N_k}} {{\sigma ^2}}\le \frac{1}{{{N_k}}}{\sigma ^2},
\end{align*}
the penultimate inequality arises from Assumption \ref{ass_1}, we can deduce that the second term of the third equation becomes zero. Then, the lemma is proved.

In general, we replace the gradient matrix ${\nabla F_k}$ in (\ref{eq:mgda_1}) by a general matrix ${W = \left[ {{w_1}, \cdots ,{w_M}} \right] \in {\mathbb{R}^{M \times N}}}$ obtain
\begin{align}
d\left( W \right) = \mathop {\arg \min }\limits_{d \in {\mathbb{R}^N}} \mathop {\max }\limits_{i \in \left[ M \right]} \{ (w^i)^Td\}  + \frac{1}{2}\left\| d \right\|^2. \label{eq:dw}
\end{align}
From (\ref{eq:mgda_2_2}) and (\ref{eq:mgda_2_1}) we have
\begin{align}
\lambda \left( W \right) = \mathop {\arg \min }\limits_{\lambda  \in {\Delta _M}} \left\| {\sum\limits_{i = 1}^M {{\lambda ^i}{w^i}} } \right\|^2, \label{eq:dw_}
\end{align}
and
\begin{align}
d\left( W \right) &= \sum\limits_{i = 1}^M {\lambda {{\left( W \right)}^i}{w^i}}. \label{eq:dw_}
\end{align}
Then we analyze the continuity properties of optimal solution ${d\left( W \right)}$ with respect to ${W}$ in (\ref{eq:dw_}).
\begin{lemma}
\label{lemma_2}
Suppose that Assumptions \ref{ass_1} and \ref{ass_2} hold. If ${W,V \in {\mathbb{R}^{M \times N}}}$ are both bounded by ${C>0}$,
then the mapping in (\ref{eq:dw}) is Hölder continuous with exponent 1/2, i.e.,
\[\left\| {d\left( W \right) - d\left( V \right)} \right\| \le \sqrt {2C} {\left\| {W - V} \right\|^{\frac{1}{2}}}.\]
\end{lemma}
\emph{proof:}
For all ${d \in {\mathbb{R}^N}}$, consider
\begin{align*}
{\phi _W}(d) &= \mathop {\max }\limits_{i \in \left[ M \right]} \{ (w^i)^Td\},\\
{\phi _V}(d) &= \mathop {\max }\limits_{i \in \left[ M \right]} \{ (v^i)^Td\},
\end{align*}
where ${W,V \in {\mathbb{R}^{M \times N}}}$. We can observe that ${{\phi _W}(d)}$ and ${{\phi _V}(d)}$ are convex sublinear functional. It follows that
  \begin{align}\label{eq:holder_0}
\left| {{\phi _W}(d) - {\phi _V}(d)} \right| \le \left\| {W - V} \right\|\left\| d \right\|.
\end{align}
By the definition of ${{d\left( W \right)}}$ and ${{{\phi _W}(d)}}$, we can rewrite (\ref{eq:dw}) as
\[d\left( W \right) = \mathop {\arg \min }\limits_{d \in {\mathbb{R}^N}} {\phi _W}(d) + \frac{1}{2}{\left\| d \right\|^2}.\]
Since the objective function of the above minimization problem is 1-strongly convex, we have
\begin{align}\label{eq:holder_1}
{\phi _W}(d) + \frac{1}{2}{\left\| d \right\|^2} \ge {\phi _W}(d\left( W \right)) + \frac{1}{2}{\left\| {d\left( W \right)} \right\|^2} + \frac{1}{2}{\left\| {d\left( W \right) - d} \right\|^2}.
\end{align}
Then
\begin{align}\label{eq:holder_2}
 {\phi _V}(d) + \frac{1}{2}\left\| d \right\|_2^2 \ge {\phi _W}(d)+\frac{1}{2}\left\| d \right\|_2^2 - \left\| {W - V} \right\|\left\| d \right\|.
\end{align}
Substituting (\ref{eq:holder_1}) into (\ref{eq:holder_2}), we obtain
\[{\phi _V}(d) + \frac{1}{2}\left\| d \right\|_2^2 \ge {\phi _W}(d\left( W \right)) + \frac{1}{2}{\left\| {d\left( W \right)} \right\|^2} + \frac{1}{2}{\left\| {d\left( W \right) - d} \right\|^2} - \left\| {W - V} \right\|\left\| d \right\|.\]
In view of the fact that ${d}$ is arbitrary, replacing $d$ in the above inequality with ${d\left( V \right)}$, we have
\begin{equation}
\begin{split} \label{eq:holder_3}
{\phi _V}(d\left( V \right)) + \frac{1}{2}\left\| {d\left( V \right)} \right\|_2^2 \ge &{\phi _W}(d\left( W \right)) + \frac{1}{2}{\left\| {d\left( W \right)} \right\|^2} + \frac{1}{2}{\left\| {d\left( W \right) - d\left( V \right)} \right\|^2}\\
 &- \left\| {W - V} \right\|\left\| {d\left( V \right)} \right\|.
\end{split}
\end{equation}
Similarly, for ${d\left( V \right)}$ we have the following inequality holds
\begin{equation}
\begin{split} \label{eq:holder_4}
{\phi _W}(d\left( W \right)) + \frac{1}{2}\left\| {d\left( W \right)} \right\|_2^2 \ge &{\phi _V}(d\left( V \right)) + \frac{1}{2}{\left\| {d\left( V \right)} \right\|^2} + \frac{1}{2}{\left\| {d\left( V \right) - d\left( W \right)} \right\|^2}\\
& - \left\| {W - V} \right\|\left\| {d\left( W \right)} \right\|.
\end{split}
\end{equation}
Combining (\ref{eq:holder_3}) and (\ref{eq:holder_4}), and rearranging the inequality, we have
\begin{align*}
{\left\| {d\left( V \right) - d\left( W \right)} \right\|^2} &\le \left\| {W - V} \right\|\left\| {d\left( W \right)} \right\| + \left\| {W - V} \right\|\left\| {d\left( V \right)} \right\|\\
& \le \left\| {W - V} \right\|\left( {\left\| {d\left( W \right)} \right\| + \left\| {d\left( V \right)} \right\|} \right)\\
& \le \left\| {W - V} \right\|\left( {\left\| {\sum\limits_{i = 1}^M {\lambda {{\left( W \right)}^i}{w^i}} } \right\| + \left\| {\sum\limits_{i = 1}^M {\lambda {{\left( V \right)}^i}{v^i}} } \right\|} \right)\\
& \le 2C\left\| {W - V} \right\|,
\end{align*}
where the last inequality holds due to the fact that ${{\lambda  \in {\Delta _M}}}$ for all ${i \in \left[ M \right]}$ and the boundedness of both $W$ and ${V}$. Thus, it follows that for all ${{ W,V \in {\mathbb{R}^{M \times N}}}}$ we have
\[\left\| {d\left( W \right) - d\left( V \right)} \right\| \le \sqrt {2C} {\left\| {W - V} \right\|^{\frac{1}{2}}}.\]
The lemma is proved.

On this basis we will further establish the convergence of iterative sequences generated by DSSMG method converge to a Pareto critical point.

\begin{theorem}\label{th:msmga}
   Suppose that Assumptions \ref{ass_1} and \ref{ass_2} hold. We set step size ${{{\alpha _k} = \frac{1}{k}}}$. For all $q>0$ and ${N_B>0}$, the sequence ${{x_1}, \cdots ,{x_K}}$ generated by DSSMG method with dynamic sample size ${{N_k} = \max \left\{ {{N_B},{k^q}} \right\}}$. Let ${\nabla {F_k} = \nabla F\left( {{x_k}} \right)}$, and ${\lambda _k^*}$ obtained from full gradient at point $x_k$ for ${k \in \left[ K \right]}$, i.e. ${\lambda _k^* = \mathop {\arg \min }\limits_{\lambda  \in {\Delta _M}} {\left\| {\nabla {F_k}\lambda } \right\|^2}}$. If the approximate gradient ${\left\{ {{Y_k}} \right\}_{k = 1}^K}$ are all bounded, then
\[\mathop {\lim }\limits_{K \to \infty } \mathbb{E}\left[ {\sum\limits_{k = 1}^K {{\alpha _k}{{\left\| {\nabla {F_k}\lambda _k^*} \right\|}^2}} } \right] < \infty. \]
\end{theorem}
\emph{proof:}
Since ${\nabla {f^i}}$, ${i \in \left[ M \right]}$ are lipschitz continuity with constant $L$, we have
\begin{align}\label{eq:descent_lemma}
{f^i}({x_{k + 1}}) = {f^i}({x_k} - {\alpha _k}{d_k}) \le {f^i}({x_k}) - {\alpha _k}\nabla {f^i}{({x_k})^T}{d_k} + \frac{L}{2}{\left\| {{\alpha _k}{d_k}} \right\|^2},
\end{align}
At ${k}$th iteration we have
\begin{equation}\label{eq:th_1}
\begin{split}
{f^i}({x_{k + 1}}) &= {f^i}({x_{k + 1}}) - {f^i}({x_k}) + {f^i}({x_k})\\
& \le \mathop {\max }\limits_{i \in \left[ M \right]} \{ {f^i}({x_{k + 1}}) - {f^i}({x_k})\}  + {f^i}({x_k})\\
& \le \mathop {\max }\limits_{i \in \left[ M \right]} \{ \nabla {f_i}{({x_k})^T}( - {\alpha _k}{Y_k}{\lambda _k}) + \frac{L}{2}\left\| {{\alpha _k}{Y_k}{\lambda _k}} \right\|_2^2\}  + {f^i}({x_k})\\
& = \mathop {\max }\limits_{i \in \left[ M \right]} \{ \nabla {f^i}{({x_k})^T}( - {\alpha _k}{Y_k}{\lambda _k} + {\alpha _k}\nabla {F_k}\lambda _k^* - {\alpha _k}\nabla {F_k}\lambda _k^*) + \frac{L}{2}\left\| {{\alpha _k}{Y_k}{\lambda _k}} \right\|_2^2\}  + {f^i}({x_k})\\
& \le {\alpha _k}\mathop {\max }\limits_{i \in \left[ M \right]} \{ \nabla {f^i}{({x_k})^T}\left( { - \nabla {F_k}\lambda _k^*} \right)\}  + \mathop {\max }\limits_{i \in \left[ M \right]} \{ \nabla {f^i}{({x_k})^T}({\alpha _k}\nabla {F_k}\lambda _k^* - {\alpha _k}{Y_k}{\lambda _k})\} \\
&~~~ + \frac{L}{2}\left\| {{\alpha _k}{Y_k}{\lambda _k}} \right\|_2^2 + {f^i}({x_k}).
\end{split}
\end{equation}
Here, the second inequality follows from (\ref{eq:descent_lemma}).
Given that both ${\nabla F_k}$ and ${Y_k}$ are bounded, there exist $C>0$ such that
${\left\| {\nabla {F_k}} \right\| \le C }$ and ${\left\| {{Y_k}} \right\| \le C}$. It follows from Lemma~\ref{lemma_2} that
\begin{align}\label{eq:th_lemma}
  \left\| {\nabla {F_k}\lambda _k^* - {Y_k}{\lambda _k}} \right\| \le \sqrt {2C} {\left\| {\nabla {F_k} - {Y_k}} \right\|^{\frac{1}{2}}}.
\end{align}
Note that for selected ${q>0}$ we have ${{N_B} \ge {k^q}}$. Taking both sides of (\ref{eq:th_lemma}) to fourth power and conditional expectation of $k$th step, it follows that
\begin{equation}\label{eq:th_2}
\begin{split}
{\mathbb{E}_k}\left[ {{{\left\| {\nabla {F_k}\lambda _k^* - {Y_k}{\lambda _k}} \right\|}^4}} \right] &\le {\mathbb{E}_k}\left[ {{{\left( {2C} \right)}^2}{{\left\| {\nabla {F_k} - {Y_k}} \right\|}^2}} \right]\\
& \le {\left( {2C} \right)^2}\frac{1}{{{N_B}}}{\sigma ^2}\\
& \le {\left( {2C\sigma } \right)^2}{\left( {\frac{1}{k}} \right)^q},
\end{split}
\end{equation}
where the second inequality can be obtained directly from Lemma~\ref{lemma_2}. Now consider second term of the last inequality in (\ref{eq:th_1}), for all ${\gamma ,\eta  > 0}$, we have
\begin{equation}\label{eq:th_2}
\begin{split}
\mathop {\max }\limits_{i \in \left[ M \right]} \left\{ {\nabla {f^i}{{({x_k})}^T}\left( {{\alpha _k}\nabla {F_k}\lambda _k^* - {\alpha _k}{Y_k}{\lambda _k}} \right)} \right\} &\le \frac{1}{2}\gamma {\left( {{\alpha _k}} \right)^2} + \frac{1}{{2\gamma }}C^2{\left\| {\nabla {F_k}\lambda _k^* - {Y_k}{\lambda _k}} \right\|^2}\\
&\le \frac{1}{2}\gamma {\left( {{\alpha _k}} \right)^2} + \frac{1}{2}\eta {\left( {\frac{1}{{2\gamma }}C^2} \right)^2} + \frac{1}{{2\eta }}{\left\| {\nabla {F_k}\lambda _k^* - {Y_k}{\lambda _k}} \right\|^4},
\end{split}
\end{equation}
both of the above inequalities follow from the young's inequality. Then repeating the same operation on both sides of the inequality yields
\begin{equation}\label{eq:th_3}
\begin{split}
&{E_k}\left[ {\mathop {\max }\limits_{i \in \left[ M \right]} \left\{ {\nabla {f^i}{{({x_k})}^T}\left( {{\alpha _k}\nabla {F_k}\lambda _k^* - {\alpha _k}{Y_k}{\lambda _k}} \right)} \right\}} \right] \\
&\le \frac{1}{2}\gamma {\left( {{\alpha _k}} \right)^2} + \frac{1}{2}\eta {\left( {\frac{1}{{2\gamma }}C^2} \right)^2} + \frac{1}{{2\eta }}{\left( {2C\sigma } \right)^2}{\left( {\frac{1}{k}} \right)^q}\\
& \le \frac{1}{2}\gamma {\left( {{\alpha _k}} \right)^2} + \frac{1}{{\gamma }}C^3\sigma {\left( {\frac{1}{k}} \right)^{\frac{q}{2}}}\\
& \le {\alpha _k}\sqrt {2C^3\sigma } {\left( {\frac{1}{k}} \right)^{\frac{q}{4}}},
\end{split}
\end{equation}
where the first inequality follows from (\ref{eq:th_2}). Let ${\eta_k  = \frac{{4\gamma \sigma_k }}{C}{\left( {\frac{1}{k}} \right)^{\frac{q}{2}}}}$ and ${\gamma_k  = \frac{{\sqrt {2{C^3}\sigma_k } }}{{{\alpha _k}}}{\left( {\frac{1}{k}} \right)^{\frac{q}{4}}}}$. Taking expectation of $k$th step of (\ref{eq:th_1}) and plugging in (\ref{eq:th_3}) into we have
 \begin{equation}\label{eq:th_4}
\begin{split}
{\mathbb{E}_k}\left[ {{f^i}\left( {{x_{k + 1}}} \right)} \right] &\le {\alpha _k}{\mathbb{E}_k}\left[ {\mathop {\max }\limits_{i \in \left[ M \right]} \left\{ {\nabla {f^i}{{\left( {{x_k}} \right)}^T}\left( { - \nabla {F_k}\lambda _k^*} \right)} \right\}} \right] + {\mathbb{E}_k}\left[ {\mathop {\max }\limits_{i \in \left[ M \right]} \{ \nabla {f^i}{{({x_k})}^T}({\alpha _k}\nabla {F_k}\lambda _k^* - {\alpha _k}{Y_k}{\lambda _k})\} } \right]\\
& ~~+ {\mathbb{E}_k}\left[ {\frac{L}{2}\left\| {{\alpha _k}{Y_k}{\lambda _k}} \right\|^2} \right] + {\mathbb{E}_k}\left[ {{f^i}({x_k})} \right]\\
 &\le  - {\alpha _k}\left\| {\nabla {F_k}\lambda _k^*} \right\|^2 + {\alpha _k}\sqrt {2C^3\sigma }{\left( {\frac{1}{k}} \right)^{\frac{q}{4}}} + \frac{L}{2}\alpha _k^2{C^2}+ {\mathbb{E}_k}\left[ {{f^i}({x_k})} \right],
\end{split}
 \end{equation}
where the first term of the last inequality directly from Lemma 4.1 in \cite{tanabe2019proximal} with ${\ell  = 1}$. Summing inequality (\ref{eq:th_4}) over ${k = 0, \cdots ,K}$,  and taking total expectation we obtain that
 \begin{align*}
\sum\limits_{k = 1}^K {\frac{1}{k}} \mathbb{E}\left[ {{{\left\| {\nabla {F_k}\lambda _k^*} \right\|}^2}} \right] &\le \sqrt {2{C^3}\sigma } \sum\limits_{k = 1}^K {{{\left( {\frac{1}{k}} \right)}^{1 + \frac{q}{4}}}}  + \frac{L}{2}{C^2}\sum\limits_{k = 1}^K {{{\left( {\frac{1}{k}} \right)}^2}}
 + \sum\limits_{k = 1}^K {\mathbb{E}\left[ {{f^i}({x_k})} \right]}  - \sum\limits_{k = 1}^K {\mathbb{E}\left[ {{f^i}({x_{k + 1}})} \right]} \\
& \le \sqrt {2{C^3}\sigma } \sum\limits_{k = 1}^K {{{\left( {\frac{1}{k}} \right)}^{1 + \frac{q}{4}}}} + \frac{L}{2}{C^2}\sum\limits_{k = 1}^K {{{\left( {\frac{1}{k}} \right)}^2}}  + {f^i}({x_1}) - \mathbb{E}\left[ {{f^i}({x_{K + 1}})} \right]\\
& \le \sqrt {2{C^3}\sigma } \sum\limits_{k = 1}^K {{{\left( {\frac{1}{k}} \right)}^{1 + \frac{q}{4}}}}  + \frac{L}{2}{C^2}\sum\limits_{k = 1}^K {{{\left( {\frac{1}{k}} \right)}^2}}  + {F_0} - {F_{\inf }},
 \end{align*}
 where ${{F_0} = \mathop {\max }\limits_{i \in \left[ M \right]} \left\{ {{f^i}({x_1})} \right\}}$ and the last inequality follows from Assumption \ref{ass_2}(a). Since ${1+{\frac{q}{4}>1}}$, taking ${K \to \infty }$, we obtain ${\sum\limits_{k = 1}^\infty  {{{\left( {\frac{1}{k}} \right)}^{1 + \frac{q}{4}}}}  < \infty }$. Thus
 \[\mathop {\lim }\limits_{K \to \infty } \mathbb{E}\left[ {\sum\limits_{k = 1}^K {{\alpha _k}{{\left\| {\nabla {F_k}\lambda _k^*} \right\|}^2}} } \right] < \infty. \]
The theorem is proved.

Theorem \ref{th:msmga} indicates that the direction ${ {\nabla {F_k}\lambda _k^*}}$ of the full gradient method converges to 0, implying that the iterative sequence $x_k$, ${{k \in \left[ K\right]}}$ generated by the DSSMG method can converge to a Pareto critical point. Therefore, we present a MOO stochastic gradient method with convergence, based on this we can design security guarding methods for multi objective situations.

\subsection{GML2O}

\begin{figure}[h!]
\centering
\includegraphics[scale=0.4]{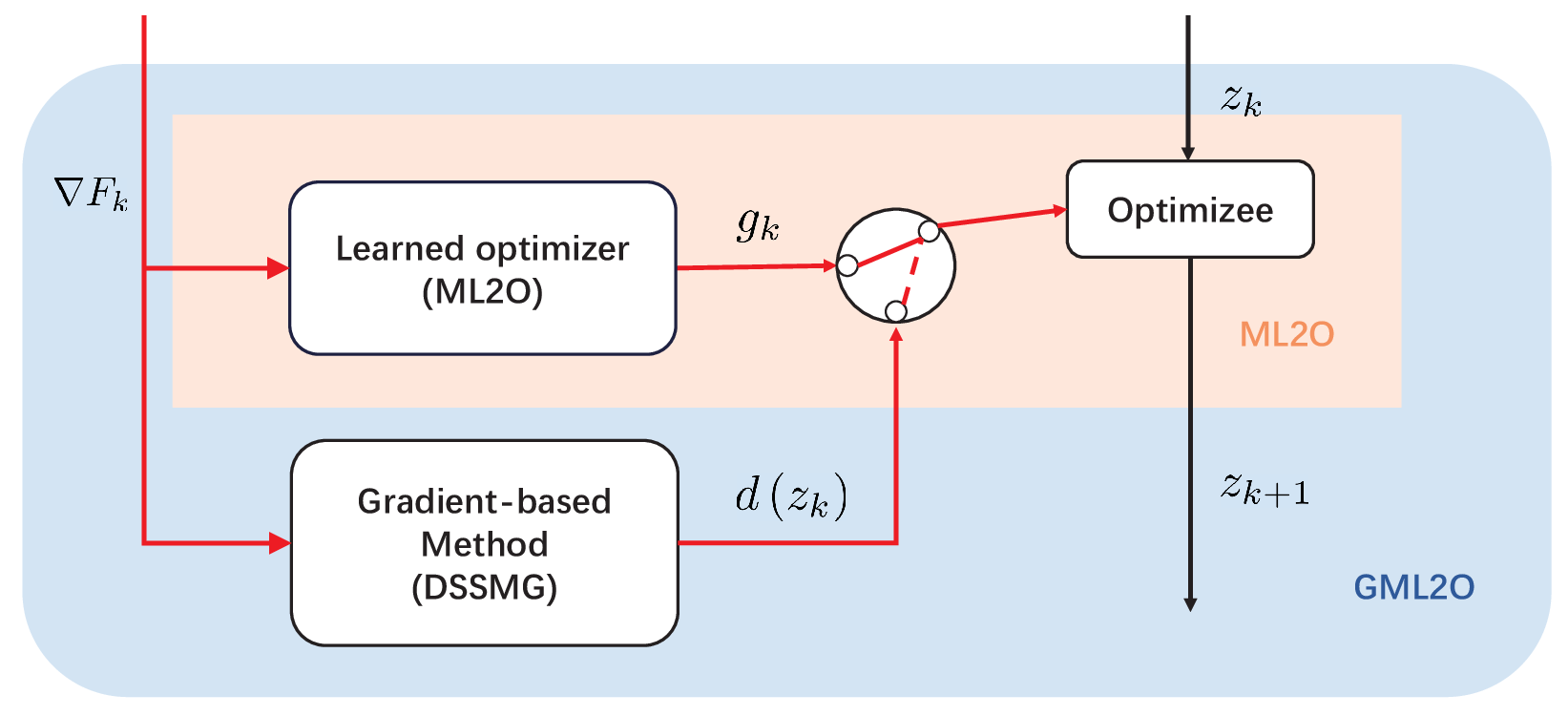}
\caption{Diagram of GML2O. GML2O adds a guarding mechanism to the learned optimizer (e.g., ML2O) that monitors for underperformance of the learned optimizer and switches to an analytical method (e.g., DSSMG) in these cases. }
\label{fig:safe_ml2o}
\end{figure}

In GML2O, the guarding mechanism evaluates the update proposed by ML2O and decides whether to accept them or substitute by update suggested by the convergence guaranteed optimizer, this substitution is called as fallback update. Principle of guarding is illustrated in Figure~\ref{fig:safe_ml2o}.

In contrast to existing single-objective guarding criterion \cite{premont2022simple,heaton2020safeguarded}, the design of MOO guarding criterion necessitates the consideration of multiple objectives. We introduce an MOO version of the guarding criterion to ensure the convergence of GML2O by employing the proven convergent DSSMG method as a fallback update.
In every $k$th iteration, update proposed by ML2O is denoted as $u_k$, while the update of DSSMG is denoted as $x_k$. In the same step size setting, if the maximum difference in objective function values between two consecutive steps, as determined by ML2O, is less than the maximum difference obtained by the gradient-based algorithm, then the update suggested by ML2O will be accepted. Conversely, the update proposed by ML2O is rejected, and the update from the algorithm with guaranteed convergence is utilized. Assuming that the sequence ${z_1}, \cdots ,{z_K}$, generated by GML2O satisfies
\begin{align}\label{eq:safe_guard}
\mathop {\max }\limits_{i \in \left[ M \right]} \left\{ {{f^i}({z_{k + 1}}) - {f^i}({z_k})} \right\} \le \min \left\{ {\mathop {\max }\limits_{i \in \left[ M \right]} \left\{ {{f^i}({x_{k + 1}}) - {f^i}({z_k})} \right\},\mathop {\max }\limits_{i \in \left[ M \right]} \left\{ {{f^i}({u_{k + 1}}) - {f^i}({z_k})} \right\}} \right\},
\end{align}
where ${{x_{k + 1}} = {z_k} - {\alpha _k}d\left( {{z_k}} \right)}$ and ${{u_{k + 1}} = {z_k} - {\alpha_k}{g_k}}$, ${\alpha_k}$ is the step size of iteration.
We provide a concise summary of the fundamental approach employed by GML2O in Algorithm \ref{alg:gml2o}.

\begin{algorithm}[h]
\begin{normalsize}
\caption{\text {GML2O}}
\begin{algorithmic} \label{alg:gml2o}
\REQUIRE  Objective function of MOO ${\mathcal{F} = \left( {{f^1}, \cdots ,{f^M}} \right)^T}$, number of iterations $K$, step size $\alpha_k$, dynamic sampling parameter $q$ and threshold $N_B$.
\ENSURE $z_1$,${H_1}$
\STATE Given MOO L2O operator: ${\mathcal{M}}$. \COMMENT{eg: ML2O}
\STATE Given MOO L2O weights: $\Theta $. \COMMENT{Taking form meta-training}
\FOR{${k = 1, \cdots ,K}$}
\STATE ${{N_k} = \max \left\{ {{N_B},{k^q}} \right\}}$
\STATE ${\Xi _k^{{N_k}} = \left( {\xi _{k,1}, \cdots ,\xi _{k,{N_k}}} \right)}$ \COMMENT{Sample batch data}
\STATE ${{u_{k}} = {z_{k}}}$ \COMMENT{ML2O Init}
\STATE ${{x_{k}} = {z_{k}}}$ \COMMENT{DSSMG Init}
\FOR{${i = 1, \cdots ,M}$}
  \STATE ${y_k^i = \frac{1}{{{N_B}}}\sum\limits_{j = 1}^{{N_B}} {\nabla {f^i}\left( {{x_k},\xi _{k,j}} \right)} }$
\ENDFOR
  \STATE ${{Y_k} = \left( {y_k^1, \cdots ,y_k^M }\right)^T }$
  \STATE ${{g_{k}} = \mathcal{M}\left( {{Y_{k}},{H_{k }};\Theta } \right)}$
  \STATE ${{u_{k+1}} = {u_{k}} - {\alpha _{k}}{g_{k}}}$ \COMMENT{ML2O Update}
  \STATE ${{d_k} = \mathop {\arg \min }\limits_{d \in {\mathbb{R}^N}} \mathop {\max }\limits_{i \in [M]} \left\{ {{{\left( {u_{k}^i} \right)}^T}d} \right\} + \frac{1}{2}{\left\| d \right\|^2}}$  \COMMENT{Solved by (\ref{eq:msmoo_lamda}) and (\ref{eq:msmoo_dk}) }
 \STATE ${{x_{k+1}} = {x_{k}} - {\alpha _{k}}{d_{k}}}$ \COMMENT{DSSMG Update}
 \IF {${\mathop {\max }\limits_{i \in \left[ M \right]} \left\{ {{f^i}({x_{k + 1}}) - {f^i}({x_k})} \right\} \le \mathop {\max }\limits_{i \in \left[ M \right]} \left\{ {{f^i}({u_{k + 1}}) - {f^i}({u_k})} \right\}}$}
 \STATE ${{z_{k+1}} = {x_{k+1}}}$
 \ELSE
 \STATE ${{z_{k+1}} = {u_{k+1}}}$
 \ENDIF
\ENDFOR
\end{algorithmic}
\end{normalsize}
\end{algorithm}
Based on Theorem~\ref{th:msmga}, we provide the key result of our work.
\begin{theorem}
Suppose that Assumptions \ref{ass_1} and \ref{ass_2} hold. For all $q>0$ and $N_B>0$, the sequence ${{z_1}, \cdots ,{z_K}}$ generated by Algorithm \ref{alg:gml2o} with a dynamic sample size ${{N_k} = \max \left\{ {{N_B},{k^q}} \right\}}$ and step size ${{{\alpha _k} = \frac{1}{k}}}$. Let ${\nabla F\left( {{z_k}} \right)}$ and ${\lambda _k^*}$ be the weight obtained from full gradient at point $z_k$ for ${k \in \left[ K \right]}$, i.e. ${\lambda _k^* = \mathop {\arg \min }\limits_{\lambda  \in {\Delta _M}} {\left\| {\nabla F\left( {{z_k}} \right)\lambda } \right\|^2}}$. If the approximate gradient ${{Y_k}}$ are all bounded, it achieves
\begin{align} \label{eq:th2}
\mathop {\lim }\limits_{K \to \infty } \mathbb{E}\left[ {\sum\limits_{k = 1}^K {{\alpha _k}{{\left\| {\nabla F\left( {{z_k}} \right)\lambda _k^*} \right\|}^2}} } \right] < \infty.
\end{align}
\end{theorem}
\emph{proof:}
 At ${k}$th iteration we have
\begin{equation}\label{eq:zk}
\begin{split}
{f^i}({z_{k + 1}}) &= {f^i}({z_{k + 1}}) - {f^i}({z_k}) + {f^i}({z_k})\\
& \le \mathop {\max }\limits_{i \in \left[ M \right]} \left\{ {{f^i}({z_{k + 1}}) - {f^i}({z_k})} \right\} + {f^i}({z_k})\\
& \le \mathop {\max }\limits_{i \in \left[ M \right]} \left\{ {{f^i}({x_{k + 1}}) - {f^i}({z_k})} \right\} + {f^i}({z_k})\\
 &= \mathop {\max }\limits_{i \in \left[ M \right]} \left\{ {{f^i}({x_{k + 1}}) - {f^i}({x_k})} \right\} + {f^i}({x_k}),
\end{split}
\end{equation}
where the sequence ${{x_1}, \cdots ,{x_K}}$ is generated by Algorithm \ref{alg:gml2o}, the last inequality is a direct application of guarding mechanism (\ref{eq:safe_guard}) and the last equation is due to the fact that ${{z_k} = {x_k}}$. The proof of Theorem \ref{th:msmga} can be referred to for the remaining part.

It's worth nothing that our proposed guarding criterion ensures the convergence of the learning optimization algorithm even in the scenario of full gradients. We present the deterministic algorithm in Algorithm \ref{alg:gml2o_full} and establish the proof of convergence when the MGDA is selected as the fallback update.
\begin{algorithm}[h!]
\begin{normalsize}
\caption{\text {Safeguarded ML2O with (deterministic) gradient descent}}
\begin{algorithmic}\label{alg:gml2o_full}
\REQUIRE Objective function of MOO ${F = \left( {{f^1}, \cdots ,{f^M}} \right)^T}$, number of iterations $K$, step size $\alpha_k$, batch size $N_B$.
\ENSURE $z_1$,${H_1}$
\STATE Given MOO L2O operator: ${\mathcal{M}}$.
\STATE Given MOO L2O weights: $\Theta $. \COMMENT{Taking form meta-training}
\FOR{${k = 1, \cdots ,K}$}
\STATE ${{g_{k}} = \mathcal{M}\left( {{\nabla F_{k}},{H_{k }};\Theta } \right)}$
\STATE ${{u_{k+1}} = {z_{k}} - {\alpha _{k}}{g_{k}}}$
  \STATE ${{d_k} = \mathop {\arg \min }\limits_{d \in {\mathbb{R}^N}} \mathop {\max }\limits_{i \in [M]} \left\{ {\nabla {f^i}{{({z_k})}^T}d} \right\} + \frac{1}{2}{\left\| d \right\|^2}}$  \COMMENT{Solved by (\ref{eq:mgda_2_1}) and (\ref{eq:mgda_2_2}) }
 \STATE ${{x_{k+1}} = {z_{k}} - {\alpha _{k}}{d_{k}}}$
 \IF {${\mathop {\max }\limits_{i \in \left[ M \right]} \left\{ {{f^i}({x_{k + 1}}) - {f^i}({z_k})} \right\} \le \mathop {\max }\limits_{i \in \left[ M \right]} \left\{ {{f^i}({u_{k + 1}}) - {f^i}({z_k})} \right\}}$}
 \STATE ${{z_{k+1}} = {x_{k+1}}}$
 \ELSE
 \STATE ${{z_{k+1}} = {u_{k+1}}}$
 \ENDIF
\ENDFOR
\end{algorithmic}
\end{normalsize}
\end{algorithm}

\begin{theorem}
Suppose that Assumption \ref{ass_2} holds. For the sequence ${{z_1}, \cdots ,{z_K}}$ generated by Algorithm \ref{alg:gml2o_full} with step size ${0 < \alpha  < \frac{2}{L}}$, we assume that ${{f^i}}$ ${(i \in \left[ M \right])}$ are all bounded from below. Let ${d\left( {{z_k}} \right)}$ be defined in (\ref{eq:mgda_1}), it achieves
\[\mathop {\lim }\limits_{k \to \infty } \left\| {d\left( {{z_k}} \right)} \right\| = 0.\]
\end{theorem}
\emph{proof:}
For all ${i \in \left[ M \right]}$, at $k$th iteration we have
\begin{align*}
{f^i}({z_{k + 1}}) &= {f^i}({z_{k + 1}}) - {f^i}({z_k}) + {f^i}({z_k})\\
 &\le \mathop {\max }\limits_{i \in \left[ M \right]} \left\{ {{f^i}({z_{k + 1}}) - {f^i}({z_k})} \right\} + {f^i}({z_k})\\
 &\le \mathop {\max }\limits_{i \in \left[ M \right]} \left\{ {{f^i}({x_{k + 1}}) - {f^i}({z_k})} \right\} + {f^i}({z_k})\\
& \le \mathop {\max }\limits_{i \in \left[ M \right]} \left\{ {\nabla {f^i}{{({z_k})}^T}{{d\left( {{z_k}} \right)}}} \right\} + \frac{L}{2}\alpha^2{\left\| {{d\left( {{z_k}} \right)}} \right\|^2} + {f^i}({z_k})\\
& \le  - {\alpha}{\left\| {{{d\left( {{z_k}} \right)}}} \right\|^2} + \frac{L}{2}\alpha ^2{\left\| {{{d\left( {{z_k}} \right)}}} \right\|^2} + {f^i}({z_k}),
\end{align*}
where the penultimate inequality comes from descent lemma obtained by Lipschitz continuity of ${\nabla {f^i}}$. Since ${0 < \alpha  < \frac{2}{L}}$ and ${{f^i}}$ is bounded from below, summing above inequality over ${k = 0, \cdots ,K}$, we obtain that
\begin{align*}
\sum\limits_{k = 1}^K {\alpha {{\left\| {{{d\left( {{z_k}} \right)}}} \right\|}^2}}  &\le \frac{1}{{\left( {1 - \frac{L}{2}\alpha } \right)}}\sum\limits_{k = 1}^K {\left( {{f^i}({z_k}) - {f^i}({z_{k + 1}})} \right)} \\
& \le \frac{1}{{\left( {1 - \frac{L}{2}\alpha } \right)}}\left( {{f^i}({z_1}) - {f^i}({z_{K + 1}})} \right)\\
 &\le \frac{1}{{\left( {1 - \frac{L}{2}\alpha } \right)}}\left( {{F_1} - {F_{\inf }}} \right),
\end{align*}
where ${{F_1} = \mathop {\max }\limits_{i \in \left[ M \right]} \left\{ {{f^i}({z_1})} \right\}}$. Taking ${K \to \infty }$, we obtain
\[\sum\limits_{k = 0}^\infty  {\alpha {{\left\| {d\left( {{z_k}} \right)} \right\|}^2}}  < \infty, \]
and hence ${\mathop {\lim }\limits_{k \to \infty } \left\| {d\left( {{z_k}} \right)} \right\| = 0}$.

The guarding criterion defined in (\ref{eq:safe_guard}) necessitates an additional operation for each learner update in GML2O, leading to increased time complexity compared to ML2O. However, this trade-off ensures convergence of the optimization process. Consequently, GML2O is particularly suitable when ML2O is not sufficiently trained or when there is a significant disparity between the distributions of training and testing.

\section{Numerical Experment}\label{sec:experment}
\begin{figure}[ht]
\centering
\includegraphics[scale=0.5]{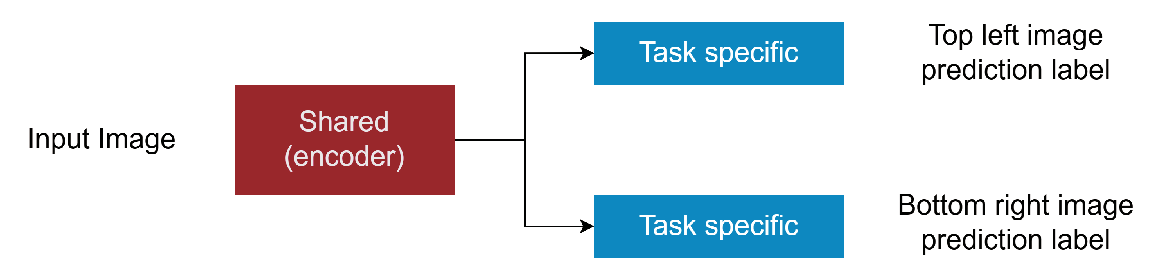}
\caption{Architecture used for MTL.}
\label{fig:hard_parameter}
\end{figure}

In this section, we present performance evaluation experiments and generalization experiments for our methods, respectively. In the performance evaluation experiment, ML2O is trained and evaluated within a consistent framework consisting of the same network structure, training settings and training dataset.
The training phase aims to acquire the optimal parameters ${\Theta}$ for the learning optimizer. Subsequently, the evaluation stage involves retraining the neural network under identical conditions while keeping the parameters of ML2O fixed at ${\Theta}$. The performance of the trained learner is then assessed on an independent test dataset.
Generalization experiments, on the other hand, train ML2O by pre-determining experimental settings, and then the trained ML2O is used as an optimizer to optimize the optimizee in different training settings, dataset or network architectures.
Similarly, a separate test dataset is used to evaluate the performance of the optimized model.
These generalization experiments aim to assess the efficacy of the learned optimizer in successfully optimizing the optimization task across different scenarios, thereby verifying its effectiveness as an optimization method.
In the specific experimental setup, we adopt a hard parameter sharing scheme for MTL, as depicted in Figure \ref{fig:hard_parameter}.

The loss function of the MTL neural network is expressed as follows
\[\mathcal{L} = {\left( {{L^1}\left( {x,\xi } \right),{L^2}\left( {x,\xi } \right)} \right)^T},\]
where $x$ is the parameter of the neural network, $\xi$ is a batch of data, and we chose ${L^1}$ and ${L^2}$ are both cross-entropy loss function.
We conducted experiments on different network architectures as follows.
\begin{itemize}
  \item \textbf{MLPs:} The activation function of MLPs is extended from sigmoid to ReLU, ELU and tanh. For the shared encoder we choose an MLPs with 4 hidden layers and for the task specific networks we choose an MLPs with 1 hidden layer, each with 50 hidden units.
  \item \textbf{CNN:} The structure of the shared encoder of the convolutional neural network is \textit{c-c-p-c-c-p}, where \textit{c} and \textit{p} represent the convolution and max-pooling, respectively. The task-specific network with structures of \textit{c-f}(CNN-1) and \textit{f-f}(CNN-2) where $f$ represent fully connected layer with 50 hidden units. Convolution kernel is with size of ${3 \times 3}$ and the max-pooling layer is with size of ${2 \times 2}$ and stride 2. Both CNN-1 and CNN-2 are trained with batch normalization.
  \item \textbf{Modified LeNet5:} We used a modified LeNet5 network with shared encoder is \textit{c-p-c-p-f-f}. Number of channels in these two convolutional layers are 10 and 20, respectively, and the size of each kernel is ${5 \times 5}$ with stride of 2. Both convolutional layers are followed by a ${2 \times 2}$ max pooling layer. The task-specific network uses a simple one-layer fully connected network with hidden units of 50. We use ReLU as the activation function in the network.
  \item \textbf{Modified VGG: } Modified VGG network with shared encoder is \textit{c-c-p-d-c-c-p-d-f-f}. The number of channels in these four convolutional layers are 32, 32, 64 and 64, where the size of each kernel is ${3 \times 3}$ and stride is 1. The fully connected layer $f$ with 50 hidden units. ReLU is used as an activation function in the network. $p$ is the max pooling layer with kernel size ${2 \times 2}$ .
\end{itemize}

We compare the performance of MOGM methods MGDA \cite{sener2018multi}, PCGrad \cite{yu2020gradient} and on optimizing the weighted sum of loss functions with traditional network optimizers stochastic gradient descent (SGD) \cite{robbins1951stochastic}, Adam, Momentum \cite{tseng1998incremental}, Adaptive Gradient (Adadelta) \cite{zeiler2012adadelta} and root mean square prop (RMSProp) \cite{tieleman2012lecture}.
All experiments are performed on a 64-bit PC with Intel(R) i5-10600KF CPU (4.10GHz) 16GB RAM GeForce RTX 3060 Ti GPU. For the traditional optimizers, we set the weighting parameter for the loss to be ${\frac{1}{M}}$, and M to be the number of tasks. For all methods we manually adjusted the learning rate and set the other parameters to the default values of Pytorch. The initial parameters for most of the learners in the experiments were sampled independently from a Gaussian distribution.
During the training process, our method was assigned a learning rate of 0.0005, whereas MGDA utilized a higher learning rate of 0.01 and PCGrad employed a slightly lower learning rate of 0.005. All other traditional optimization methods were set to use a much lower learning rate of 0.0001.
The batch size for training the optimizer is set to 32, with the exception of the DSSMG method, for which we will provide the batch size in the additional description.
For network testing, we report quantitative values as the average of 10 training measurements initialized with random parameters.
\subsection{Performance evaluation}
\label{sec:Performance evaluation}
\begin{figure}[ht]
\centering
\includegraphics[scale=0.45]{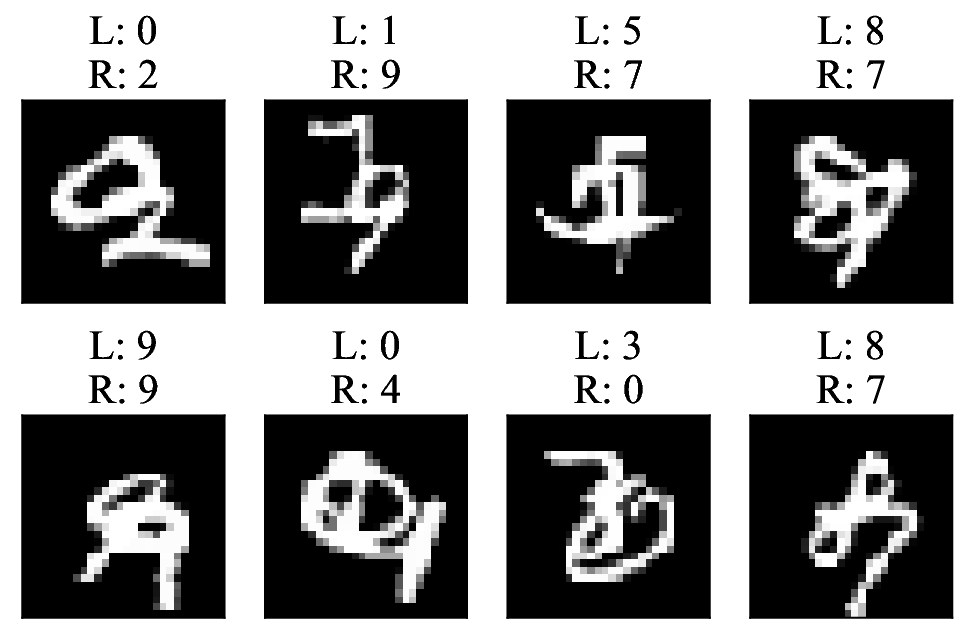}
\caption{MultiMNIST image samples. Each image is associated with two labels: one corresponding to the handwritten digit in the top left position (referred to as "L"), and the other representing the handwritten digit in the bottom right position (referred to as "R").}
\label{fig:multimnist}
\end{figure}

We evaluated our method on the MultiMNIST datasets as shown in Figure \ref{fig:multimnist}, which represent medium-sized datasets with two distinct classification tasks. To generate the MultiMNIST datasets, we followed the procedure outlined in \cite{ma2020efficient}.
Specifically, we positioned two ${28 \times 28}$ MNIST images in the top left and bottom right corners, introducing random shifts of up to 2 pixels in each direction to yield a ${36 \times 36}$ composite image. Subsequently, the synthetic images were resized to ${28 \times 28}$ and normalized with a mean of 0.1307 and a standard deviation of 0.3081. It is notable that no data augmentation techniques were employed during training or testing phases. Each dataset consisted of 60,000 training images and 10,000 test images.

\begin{figure}[ht]

\centering
\includegraphics[scale=0.3]{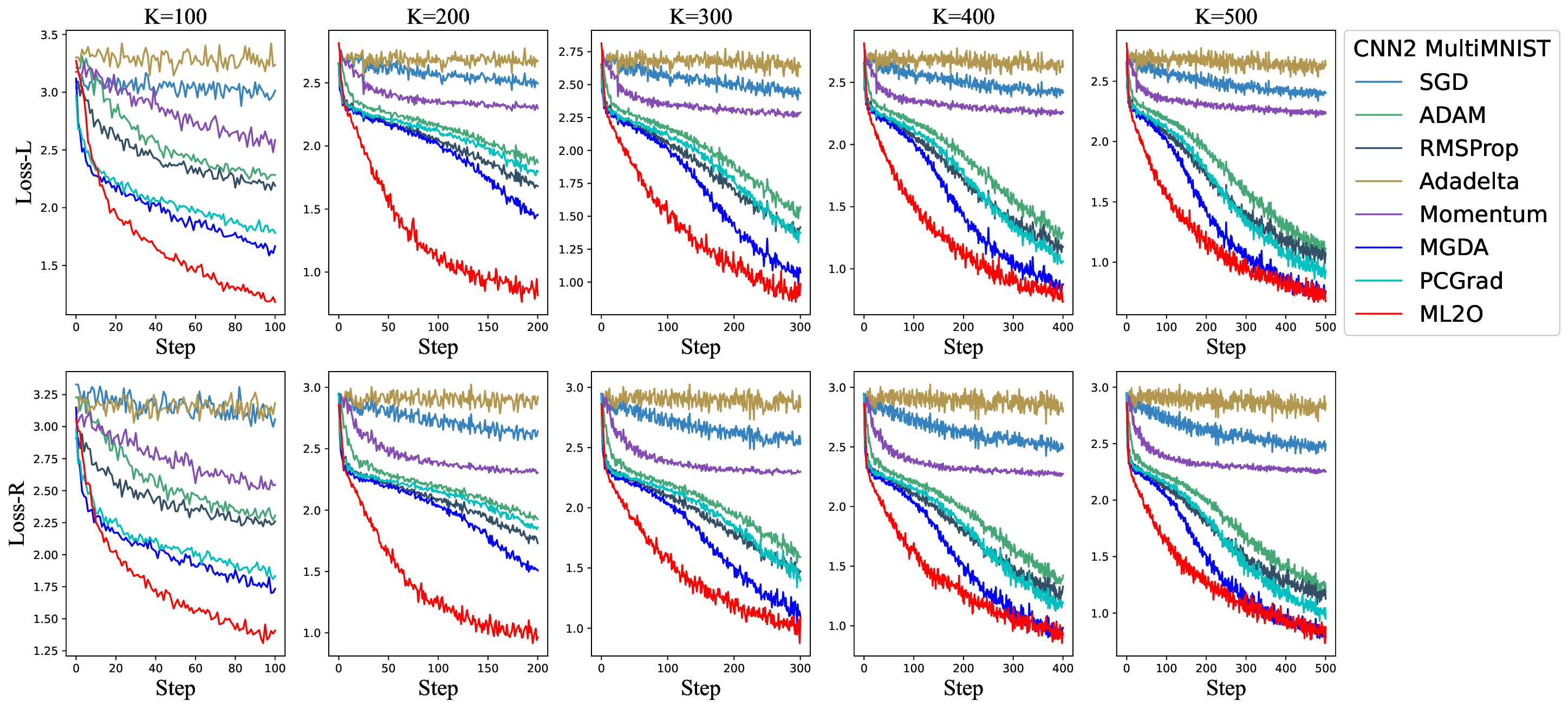}
\caption{Comparison of different optimizers for optimizing CNN2 in different optimization steps}
\label{fig:cnn2_step}
\end{figure}

\textbf{CNN} ~~We evaluate the performance of ML2O trained at different number of steps, where the learners we choose CNN1 and CNN2.
We set the maximum iterations $K = 100, 200, 300, 400, 500$, the number of candidate updates ${{\bar K}}$ to 20.
Figures \ref{fig:cnn1_step} and \ref{fig:cnn2_step} depict the loss curves for training CNN1 and CNN2 in MultiMNIST with different optimizers at different optimization steps K, respectively.
Each column in these two figures corresponds to a step size setting, and in one column the top and bottom plots illustrate the loss of the left and right plots in the classification of MultiMNIST.
It is noteworthy that our proposed method exhibits a faster decrease in loss compared to all other compared methods at optimization steps of $K = 100, 200$, and 300 for both CNN1 and CNN2. At steps 400 and 500, our method demonstrates performance comparable to MGDA and superior to the remaining six compared methods. Specifically, in the case of CNN2 and $K = 200$, our method significantly outperforms the other compared methods, with the loss in ML2O dropping below 1 while the losses of the other methods remain above 1.5.

%

\begin{figure}[ht]

\centering
\includegraphics[scale=0.3]{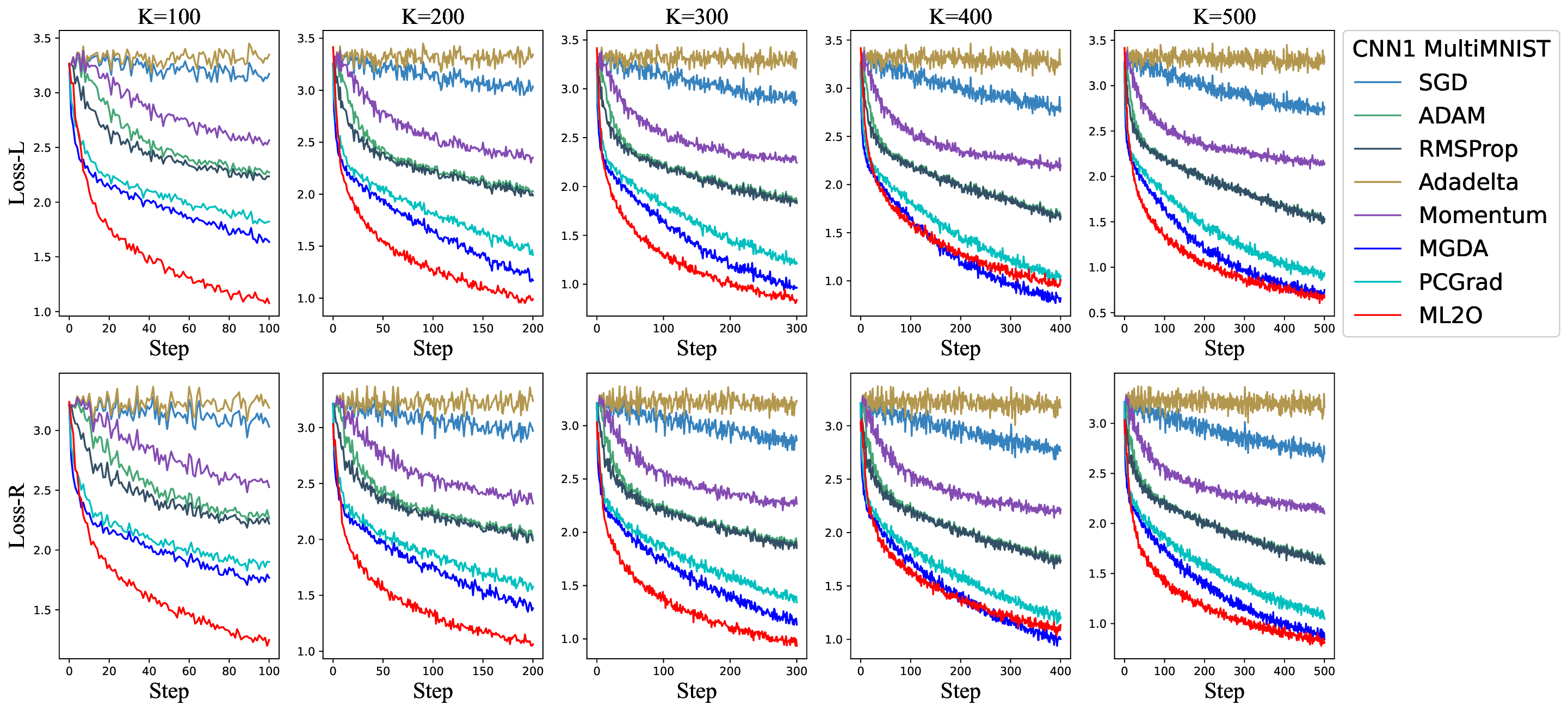}
\caption{Comparison of different optimizers for optimizing CNN1 in different optimization steps}
\label{fig:cnn1_step}
\end{figure}

\begin{figure}[h!]
\centering
\subfloat[]{\includegraphics[scale=0.3]{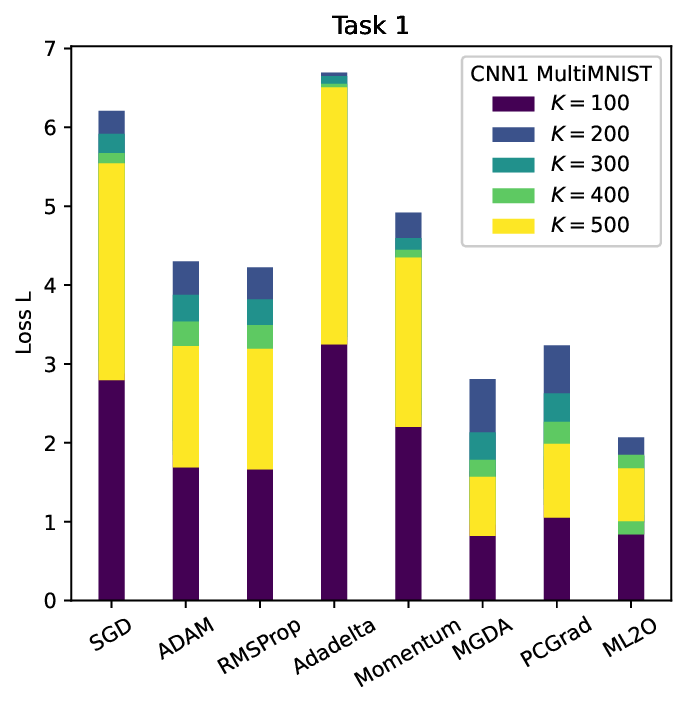}}
\subfloat[]{\includegraphics[scale=0.3]{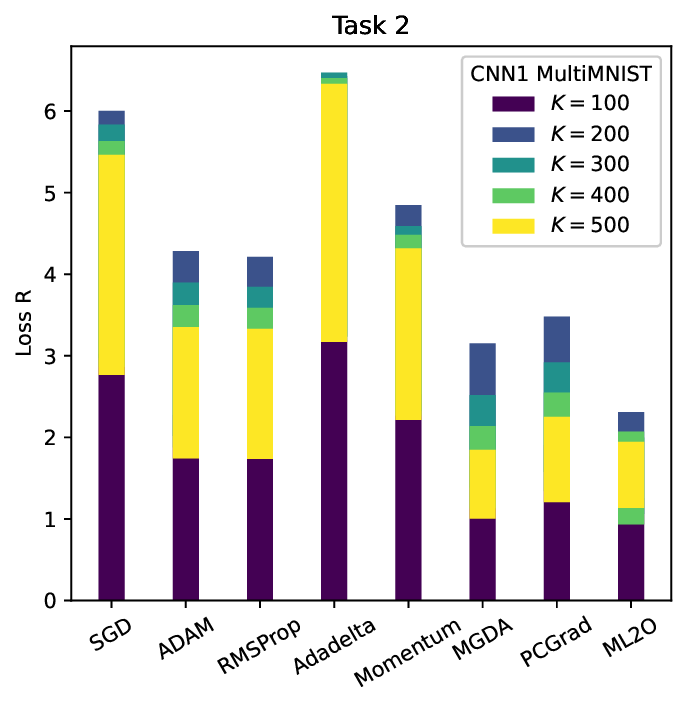}}\\
\subfloat[]{\includegraphics[scale=0.3]{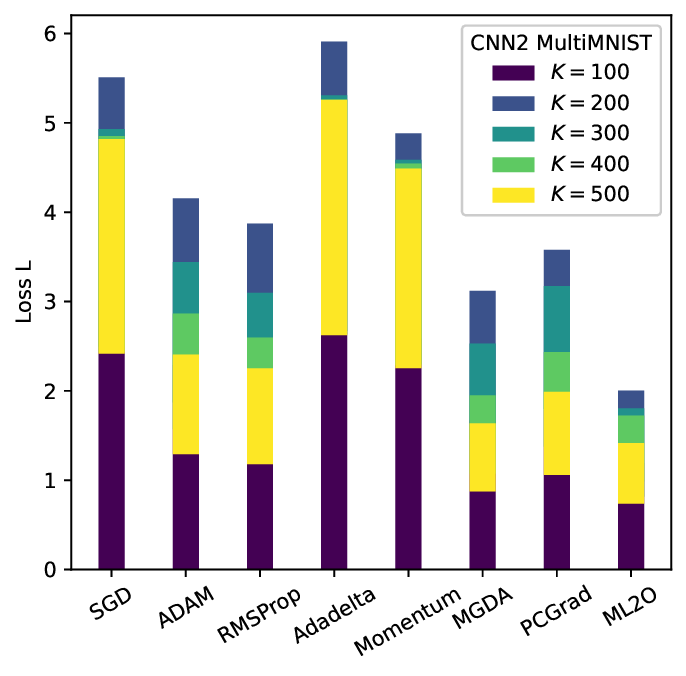}}
\subfloat[]{\includegraphics[scale=0.3]{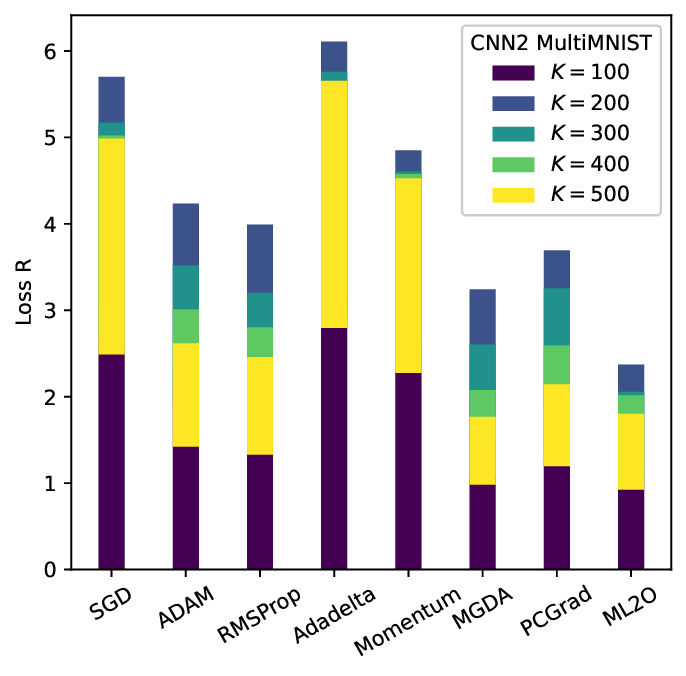}}

\caption{Comparison of the loss sum of CNN1 and CNN2 in different optimization steps and different optimizers}
\label{fig:cnn_mnist_bar}
\end{figure}

The bar chart displayed in Figure \ref{fig:cnn_mnist_bar} provides an overview of the cumulative loss of different optimizers across various optimization steps. Notably, our proposed method consistently achieves the smallest sum of losses on both CNN1 and CNN2, indicating superior performance. Hence, we can conclude that ML2O effectively reduces the training loss of convolutional neural networks and outperforms the other seven compared methods.

\begin{figure}[ht]
\centering
\includegraphics[scale=0.35]{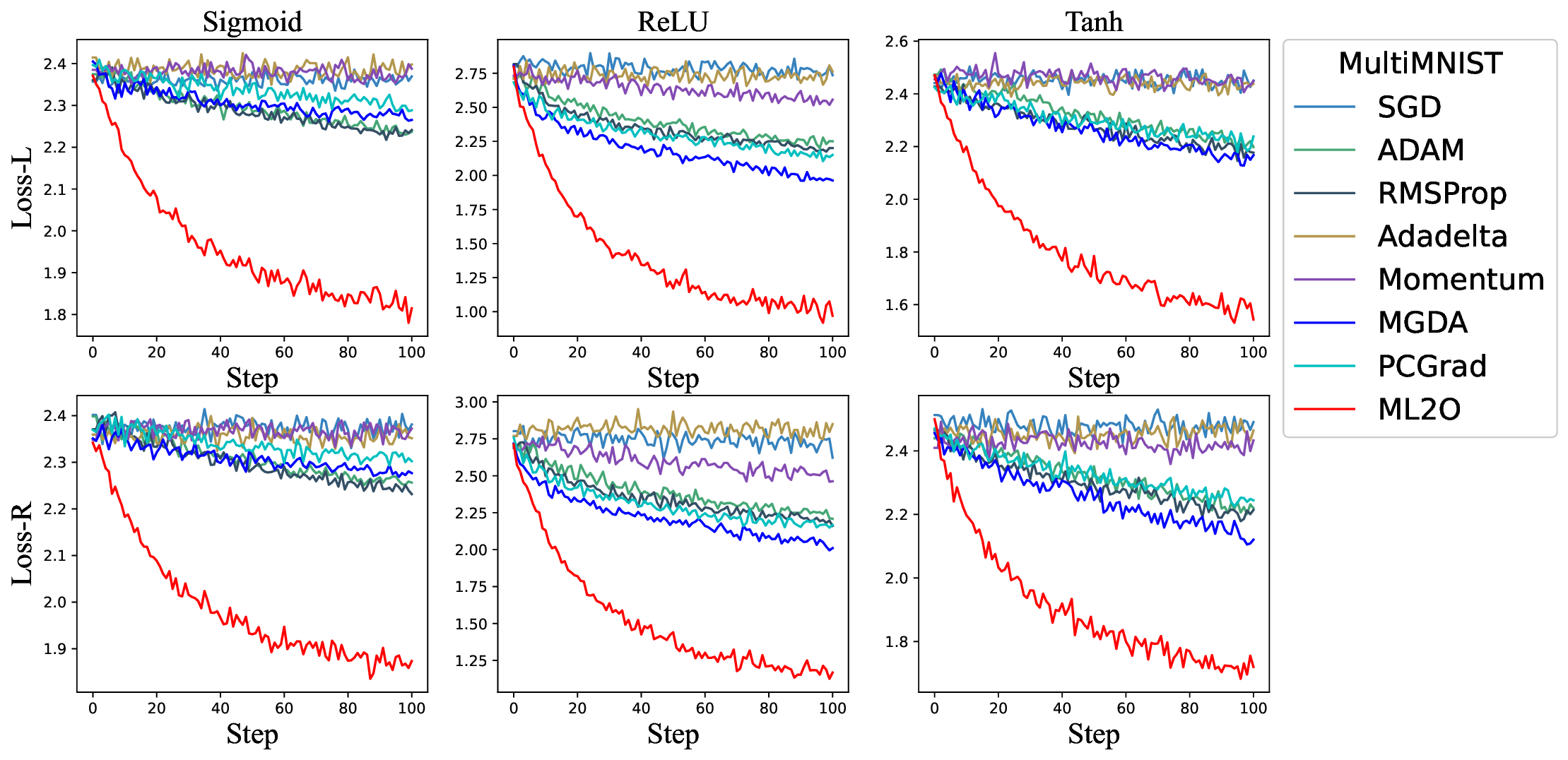}
\caption{Comparison of different optimizers for MLPs in different nonlinear activation functions}
\label{fig:mlp_mnist}
\end{figure}

\textbf{MLPs} ~~On the same dataset, we evaluated the performance of ML2O by training MLPs with different activation functions including Sigmoid, ReLU, and Tanh.
The loss curves depicting the optimization progress are presented in Figure \ref{fig:mlp_mnist}.
It is worth noting that when the Sigmoid function is used as the activation function, all seven compared methods exhibited losses above 2.2 for both tasks, while ML2O achieved losses below 1.9. Similarly, for ReLU activation, the optimal MGDA method among the comparisons yielded a loss of approximately 2.0 in the left classification loss, whereas ML2O achieved a significantly lower loss of approximately 1.0, only half that of MGDA. In the case of Tanh activation, ML2O achieved losses of approximately 1.8 (Loss-L) and 1.6 (Loss-R), demonstrating substantial improvement over the other compared methods. Thus, we can say that ML2O effectively addresses the training challenges of MLPs with different activation functions.

\subsection{Generalization}
\label{sec:Generalization}
We performed an assessment to evaluate the generalization of our  learning optimizer. This evaluation encompassed three aspects: variations in optimization steps, diverse datasets, and different network structures.
We trained ML2O on the CNN1-based MultiMNIST dataset with a fixed number of iteration steps ($K=10000$) to obtain the optimal parameter $\Theta$. Subsequently, we tested the generalization by employing ${\Theta}$ as the parameter of ML2O for optimizing various MTL problem.

\textbf{Step} ~~We conducted an evaluation to assess the generalization ability of ML2O across different optimization steps. Comparisons were made between our method and seven other optimizers at various optimization steps. Figure \ref{fig:cnn1_step} shows the loss function curves, with each column representing a different iteration step setting and each row corresponding to a distinct task. Table \ref{tab:cnn1_step} presents the performance metrics (loss, Top1 accuracy and Top5 accuracy) of the optimized learners obtained by training CNN1 with iteration steps set to $K=700, 3000, 9000, 15000$.

Observing the last three columns of Figure \ref{fig:cnn1_step}, it is evident that our method exhibits the fastest decrease in loss and reaches its optimal performance at about 2000 steps, while the other seven compared methods continue to decrease. Additionally, as indicated in Table \ref{tab:cnn1_step}, when $K=700$, the Top1 classification accuracies achieved by our optimized learners are 79.15\% and 73.11\%, significantly surpassing the 65.52\% and 55.89\% obtained by MGDA, the 45.03\% and 38.60\% achieved by PCGrad, as well as the optimal results of the traditional method, 52.83\% and 44.36\% attained by RMSProp. This superiority is attributed to the unique capability of our method to incorporate future and past information through larger optimization steps during training, leveraging the long and short term memory of the LSTM in the learning optimizer. Furthermore, as illustrated in Table \ref{tab:cnn1_step}, increasing the number of iterative steps leads to a reduction in optimizer loss and an enhancement in learner accuracy. At $K=15000$, our ML2O simultaneously classifies both tasks, yielding the highest Top1 and Top5 accuracies of 93.25\% (Loss-L) and 90.72\% (Loss-R), 99.83\% (Loss-L) and 99.76\% (Loss-R), respectively. Consequently, our ML2O demonstrates excellent performance in handling generalization tasks with varying step sizes, enabling faster training of high-quality deep neural networks by smaller optimization steps.

\begin{figure}[ht]
\centering
\includegraphics[scale=0.35]{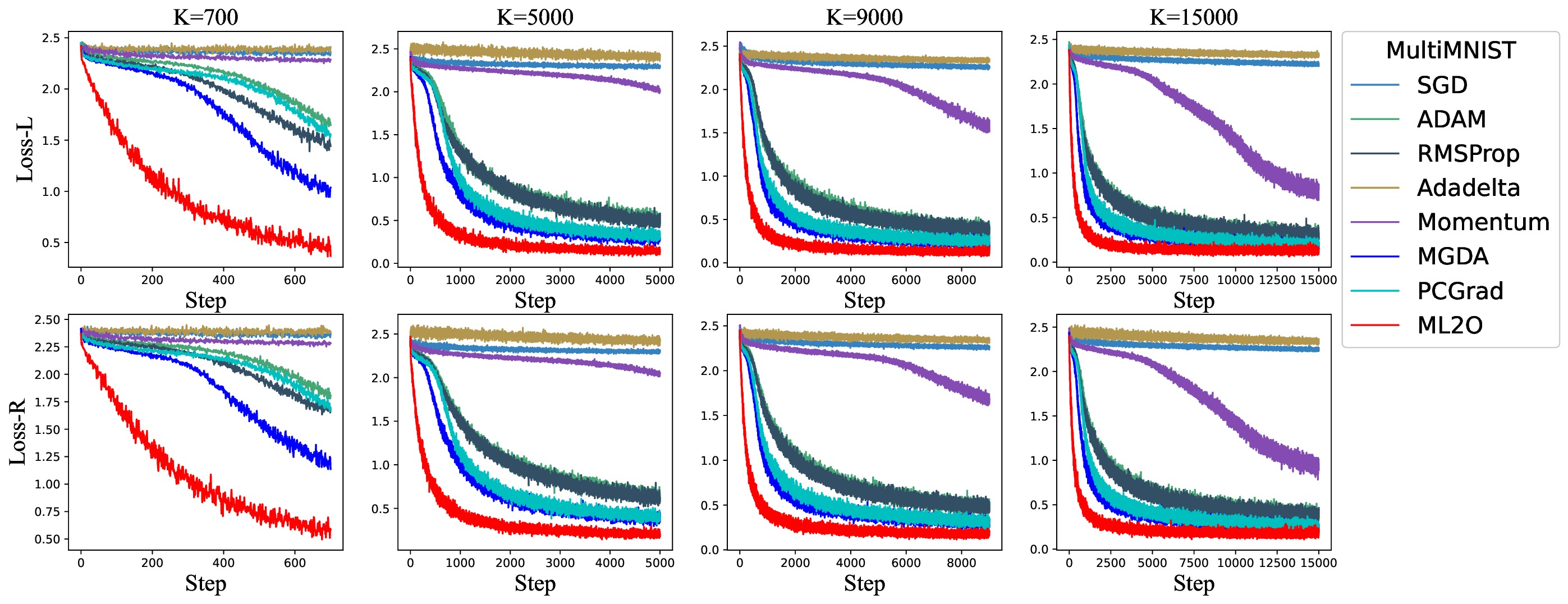}
\caption{Generalization of ML2O at optimization steps. Comparison of different optimizers for CNN1 in different optimization steps.}
\label{fig:cnn1_step}
\end{figure}

\begin{table}[h!]\footnotesize\tabcolsep 5pt
\begin{center}
\caption{Performance of CNN1 trained by different optimizers in different optimization steps}\label{tab:cnn1_step}\vspace{-2mm}
\end{center}
\begin{center}
    \begin{tabular}{r|c|c|cccccccc}
    \toprule
    \multicolumn{3}{c|}{Optimizer} & SGD   & ADAM  & Adadelta & RMSProp & Momentum & MGDA  & PCGrad & ML2O \\
    \midrule
    \multicolumn{1}{l|}{ } & \multirow{2}[2]{*}{Loss} & Task1 & 2.36  & 1.65  & 2.41  & 1.44  & 2.29  & 1.04  & 1.55  & \textbf{0.37 } \\
          &       & Task2 & 2.37  & 1.81  & 2.39  & 1.68  & 2.28  & 1.14  & 1.70  & \textbf{0.59 } \\
\cmidrule{2-11}    \multicolumn{1}{c|}{CNN1} & Top1  & Task1 & 9.73  & 45.46  & 10.00  & 52.83  & 13.64  & 62.52  & 45.03  & \textbf{79.15 } \\
    \multicolumn{1}{c|}{K=700} & Acc(\%) & Task2 & 10.10  & 38.76  & 9.74  & 44.36  & 12.26  & 55.89  & 38.06  & \textbf{73.11 } \\
\cmidrule{2-11}          & Top5  & Task1 & 50.85  & 88.14  & 51.02  & 91.57  & 55.44  & 95.32  & 86.18  & \textbf{98.04 } \\
          & Acc(\%) & Task2 & 50.05  & 84.53  & 50.07  & 88.25  & 53.96  & 92.97  & 82.07  & \textbf{97.42 } \\
    \midrule
    \multicolumn{1}{l|}{ } & \multirow{2}[2]{*}{Loss} & Task1 & 2.28  & 0.55  & 2.43  & 0.58  & 2.01  & 0.29  & 0.33  & \textbf{0.14 } \\
          &       & Task2 & 2.30  & 0.58  & 2.42  & 0.62  & 2.05  & 0.44  & 0.41  & \textbf{0.20 } \\
\cmidrule{2-11}    \multicolumn{1}{c|}{CNN1} & Top1  & Task1 & 12.17  & 83.91  & 10.33  & 84.90  & 29.98  & 88.40  & 87.87  & \textbf{90.39 } \\
    \multicolumn{1}{c|}{K=5000} & Acc(\%) & Task2 & 11.37  & 79.60  & 10.09  & 80.06  & 28.50  & 84.61  & 84.28  & \textbf{87.20 } \\
\cmidrule{2-11}          & Top5  & Task1 & 53.02  & 99.05  & 50.32  & 99.16  & 75.25  & 99.50  & 99.45  & \textbf{99.54 } \\
          & Acc(\%) & Task2 & 52.93  & 98.80  & 50.48  & 98.85  & 73.81  & 99.34  & 99.31  & \textbf{99.51 } \\
    \midrule
    \multicolumn{1}{c|}{ } & \multirow{2}[2]{*}{Loss} & Task1 & 2.26  & 0.41  & 2.35  & 0.40  & 1.52  & 0.20  & 0.29  & \textbf{0.18 } \\
          &       & Task2 & 2.26  & 0.51  & 2.35  & 0.45  & 1.68  & 0.27  & 0.33  & \textbf{0.20 } \\
\cmidrule{2-11}    \multicolumn{1}{c|}{CNN1} & Top1  & Task1 & 13.52  & 87.98  & 10.39  & 87.94  & 48.49  & 90.76  & 90.18  & \textbf{91.68 } \\
    \multicolumn{1}{c|}{K=9000} & Acc(\%) & Task2 & 13.64  & 84.42  & 10.49  & 84.51  & 42.40  & 87.74  & 87.14  & \textbf{88.61 } \\
\cmidrule{2-11}          & Top5  & Task1 & 55.40  & 99.46  & 50.28  & 99.44  & 85.83  & 99.67  & 99.61  & \textbf{99.64 } \\
          & Acc(\%) & Task2 & 55.31  & 99.33  & 51.64  & 99.31  & 83.21  & 99.59  & 99.56  & \textbf{99.63 } \\
    \midrule
    \multicolumn{1}{c|}{ } & \multirow{2}[2]{*}{Loss} & Task1 & 2.22  & 0.28  & 2.33  & 0.30  & 0.82  & 0.19  & 0.18  & \textbf{0.14 } \\
          &       & Task2 & 2.24  & 0.35  & 2.32  & 0.43  & 0.95  & 0.20  & 0.28  & \textbf{0.19 } \\
\cmidrule{2-11}    \multicolumn{1}{c|}{CNN1} & Top1  & Task1 & 17.96  & 90.55  & 10.94  & 90.53  & 75.30  & 91.76  & 92.08  & \textbf{93.25 } \\
    \multicolumn{1}{c|}{K=15000} & Acc(\%) & Task2 & 15.62  & 87.35  & 11.58  & 87.52  & 69.83  & 88.93  & 89.31  & \textbf{90.72 } \\
\cmidrule{2-11}          & Top5  & Task1 & 61.31  & 99.64  & 52.97  & 99.65  & 97.70  & 99.71  & 99.75  & \textbf{99.83 } \\
          & Acc(\%) & Task2 & 58.22  & 99.57  & 52.61  & 99.60  & 97.10  & 99.69  & 99.71  & \textbf{99.76 } \\
    \bottomrule
    \end{tabular}%
\end{center}
 \end{table}

\begin{figure}[ht]
\centering
\includegraphics[scale=0.4]{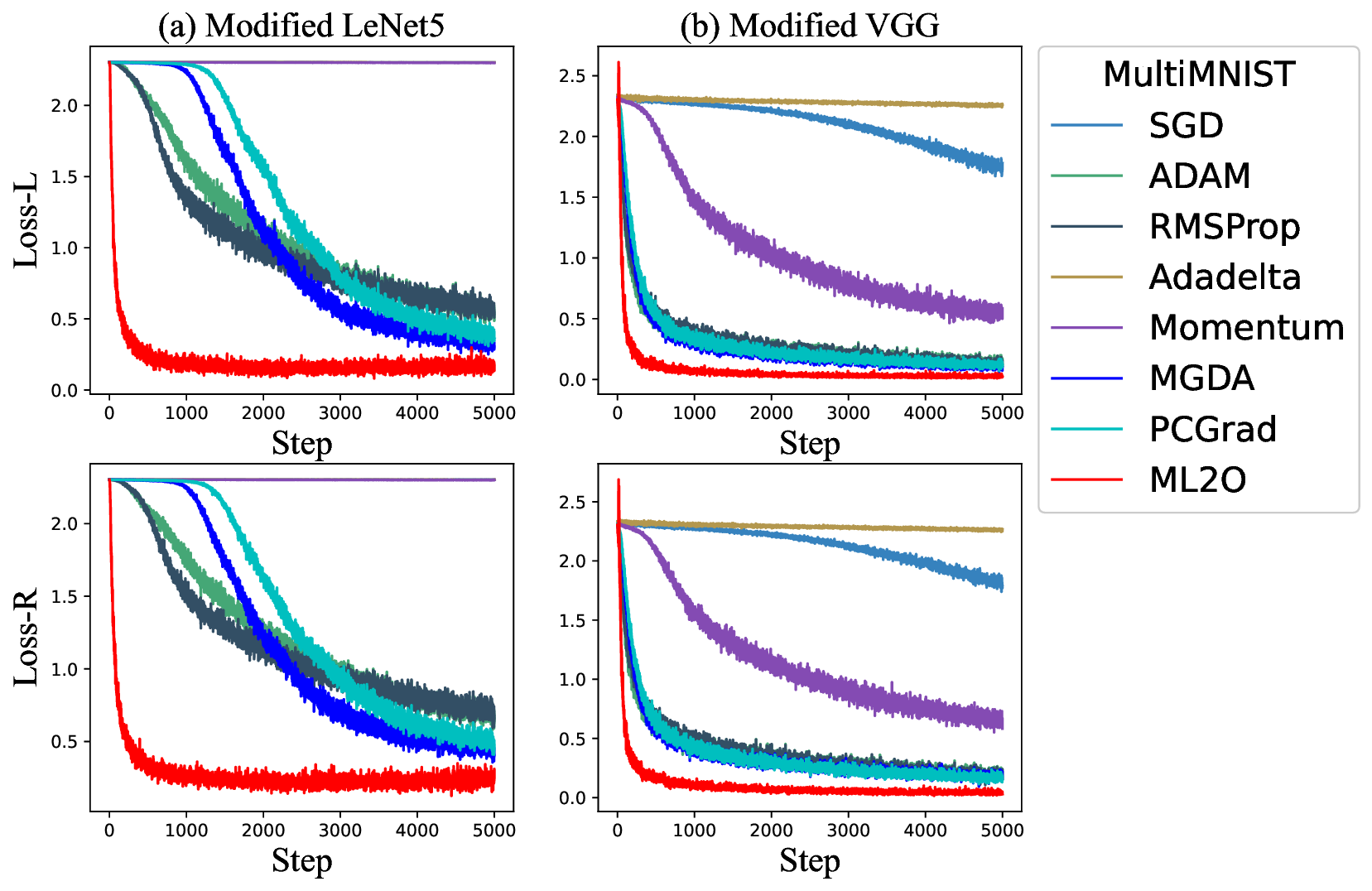}
\caption{Generalization of ML2O in terms of the network structure of learners.  Comparison of different optimizers trained in K=5000 for modified LeNet5 and modified VGG}
\label{fig:gen_mnist}
\end{figure}

\textbf{Strcture} ~~We use ML2O trained on CNN1 to optimize improved LeNet5 and improved VGG to evaluate the generalization ability of ML2O with different architectures. The decreasing trend of the loss function is depicted in Figure \ref{fig:gen_mnist}, while the corresponding loss function values, Top1 accuracy and Top5 accuracy for both tasks are reported in Table \ref{tab:gen_mnist}.

As observed in Figure \ref{fig:gen_mnist}, our method consistently achieves the highest quality solution in the shortest time compared to other methods. Particularly, when optimizing the modified LeNet5, our method attains a loss of less than 0.5 before reaching 1000 steps, whereas all other methods exhibit losses exceeding 1.0. On the other hand, when optimizing the modified VGG, our method not only demonstrates superior speed, but also exhibits exceptional stability with the smallest loss function value, which is evident from the minimal oscillation amplitude of the loss curve.
Furthermore, as shown in Table \ref{tab:gen_mnist}, our method outperforms all compared methods in terms of loss and accuracy for both tasks. We achieve the best loss and accuracy for these two learners, indicating the efficacy of ML2O in effectively reducing the loss of networks with different structures, thereby yielding superior learners. For instance, on the modified LeNet5, our loss value for Task1 is 0.14\%, which is only half of the 0.30\% achieved by MGDA. In terms of Top1 accuracy, our ML2O achieves 88.52\% for Task1 on the modified LeNet5, surpassing MGDA (87.33\%), PCGrad (85.75\%), SGD (11.55\%), ADAM (82.82\%), Adadelta (11.10\%), RMSProp (82.44\%) and Momentum (13.78\%). Similar trends can be observed for Top5 accuracy.

Consequently, we conclude that ML2O, trained on CNN1 with shallower layers and a simpler structure, can effectively serve as an optimizer for optimizing modified LeNet5 and modified VGG networks, showcasing successful generalization across different architectures. The offline-trained optimizers can be successfully employed to optimize different deep neural networks with similarly distribution, resulting in significant cost savings associated with learning ML2O.

\begin{table}[h!]\footnotesize\tabcolsep 5pt
\begin{center}
\caption{Performance of different optimizers training networks with different structures in K=5000}\label{tab:gen_mnist}\vspace{-2mm}
\end{center}
\begin{center}
    \begin{tabular}{r|c|c|cccccccc}
    \toprule
    \multicolumn{3}{c|}{Optimizer} & SGD   & ADAM  & Adadelta & RMSProp & Momentum & MGDA  & PCGrad & ML2O \\
    \midrule
    \multicolumn{1}{l|}{ } & \multirow{2}[2]{*}{Loss} & Task1 & 2.30  & 0.52  & 2.30  & 0.54  & 2.30  & 0.30  & 0.36  & \textbf{0.14 } \\
          &       & Task2 & 2.30  & 0.65  & 2.30  & 0.66  & 2.30  & 0.41  & 0.46  & \textbf{0.25 } \\
\cmidrule{2-11}    \multicolumn{1}{c|}{Modified} & Top1  & Task1 & 11.55  & 82.82  & 11.10  & 82.44  & 13.78  & 87.33  & 85.75  & \textbf{88.52 } \\
    \multicolumn{1}{c|}{LeNet5} & Acc(\%) & Task2 & 11.11  & 78.09  & 11.01  & 77.53  & 13.43  & 82.56  & 81.01  & \textbf{83.84 } \\
\cmidrule{2-11}          & Top5  & Task1 & 52.80  & 98.96  & 51.56  & 98.95  & 58.91  & 99.40  & 99.27  & \textbf{99.48 } \\
          & Acc(\%) & Task2 & 50.90  & 98.58  & 50.41  & 98.49  & 57.14  & 99.14  & 98.96  & \textbf{99.24 } \\
    \midrule
    \multicolumn{1}{c|}{ } & \multirow{2}[2]{*}{Loss} & Task1 & 1.32  & 0.45  & 2.19  & 0.44  & 0.77  & 0.28  & 0.40  & \textbf{0.23 } \\
          &       & Task2 & 1.47  & 0.52  & 2.19  & 0.51  & 0.88  & 0.48  & 0.46  & \textbf{0.19 } \\
\cmidrule{2-11}    \multicolumn{1}{c|}{Modified} & Top1  & Task1 & 48.25  & 80.67  & 23.60  & 81.00  & 67.94  & 80.14  & 79.75  & \textbf{82.86 } \\
    \multicolumn{1}{c|}{VGG} & Acc(\%) & Task2 & 44.74  & 80.59  & 21.72  & 80.83  & 67.88  & 79.75  & 79.55  & \textbf{83.19 } \\
\cmidrule{2-11}          & Top5  & Task1 & 94.96  & 99.57  & 75.39  & 99.60  & 98.81  & 99.55  & 99.54  & \textbf{99.67 } \\
          & Acc(\%) & Task2 & 93.51  & 99.56  & 73.41  & 99.60  & 98.64  & 99.57  & 99.56  & \textbf{99.69 } \\
    \bottomrule
    \end{tabular}%
\end{center}
 \end{table}

\begin{figure}[ht]
\centering
\includegraphics[scale=0.45]{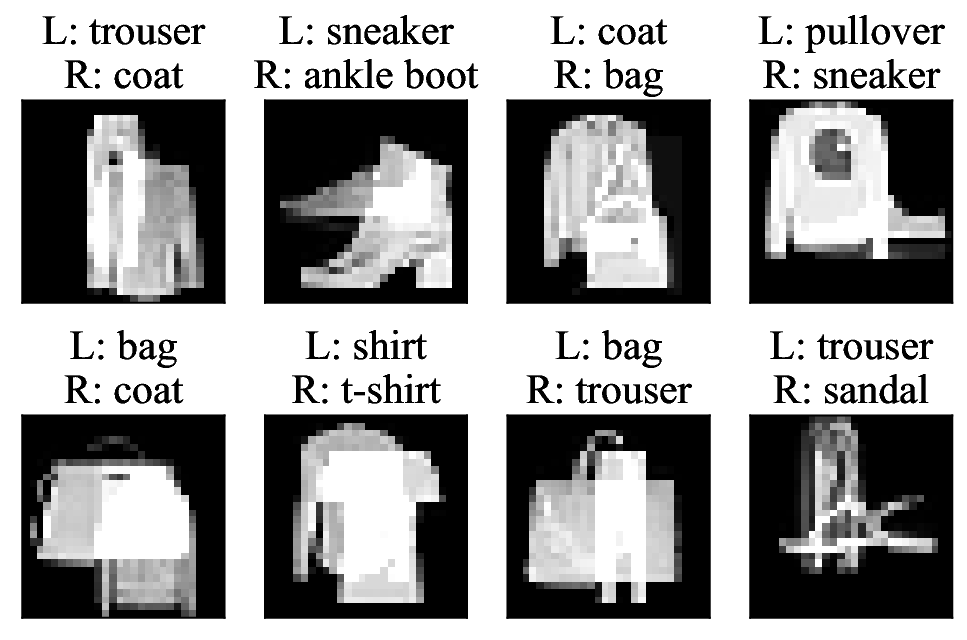}
\caption{MultiFashion image samples. Each image has two labels, one for the top left clothing image (L) and the second for the bottom right hand clothing image (R)}
\label{fig:data_fashion}
\end{figure}

\textbf{Dataset} ~~Next we evaluate the generalization ability of ML2O on different datasets. We produced a multi-task version of the MultiFashion dataset using FashionMNIST \cite{xiao2017fashion}, as shown in Figure \ref{fig:data_fashion}, and the dataset was produced with the same details as MultiMNIST.

\begin{figure}[ht]
\centering
\includegraphics[scale=0.4]{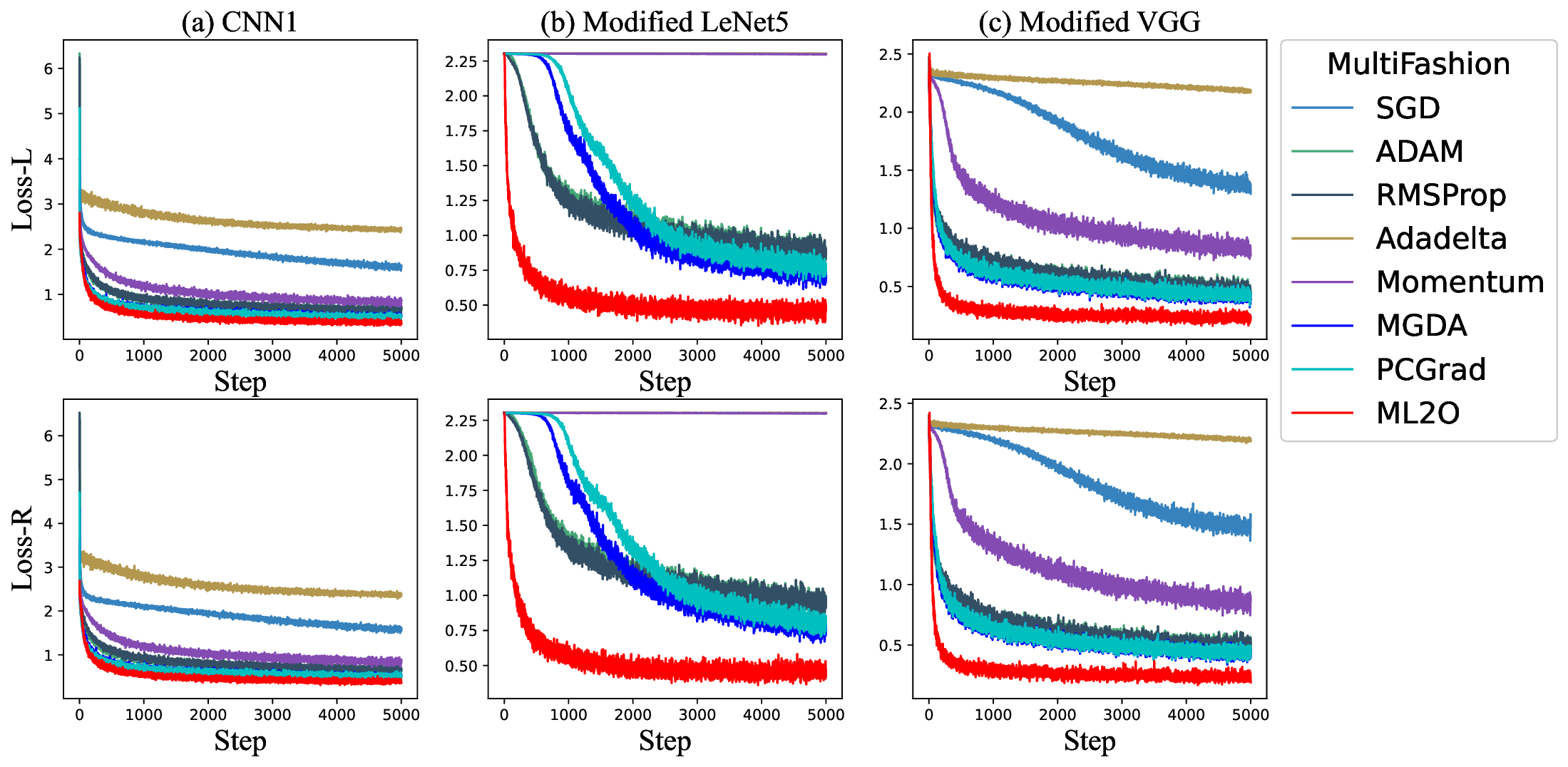}
\caption{MultiFashion image samples. Each image is associated with two labels: one corresponding to the clothing image in the top left position (referred to as "L"), and the other representing the clothing image in the bottom right position (referred to as "R").}
\label{fig:gen_fashion}
\end{figure}

We use ML2O trained on MultiMNIST to optimize the network loss on MultiFashion. According to the curves of the loss function in Figure \ref{fig:gen_fashion}, we can see that ML2O as an optimizer can effectively decrease the loss on different datasets and has outstanding generalization ability across datasets. The results in Table \ref{tab:gen_fashion} also show that the ML2O-trained learner has lower loss and higher classification accuracy on MultiFashion compared to other comparison optimizers.
\begin{table}[h!]\footnotesize\tabcolsep 5pt
\begin{center}
\caption{Performance of different optimizers to optimize MultiFashion dataset for different network structures at $K=5000$}\label{tab:gen_fashion}\vspace{-2mm}
\end{center}
\begin{center}
    \begin{tabular}{r|c|c|cccccccc}
    \toprule
    \multicolumn{3}{c|}{Optimizer} & SGD   & ADAM  & Adadelta & RMSProp & Momentum & MGDA  & PCGrad & ML2O \\
    \midrule
    \multicolumn{1}{c|}{\multirow{6}[6]{*}{CNN1}} & \multirow{2}[2]{*}{Loss} & Task1 & 1.64  & 0.71  & 2.43  & 0.73  & 0.88  & 0.55  & 0.54  & \textbf{0.36 } \\
          &       & Task2 & 1.56  & 0.69  & 2.38  & 0.67  & 0.86  & 0.54  & 0.53  & \textbf{0.36 } \\
\cmidrule{2-11}          & Top1  & Task1 & 41.40  & 74.68  & 9.64  & 74.57  & 68.66  & 75.35  & 76.59  & \textbf{79.37 } \\
          & Acc(\%) & Task2 & 43.04  & 75.28  & 13.84  & 75.04  & 69.22  & 75.95  & 76.84  & \textbf{79.50 } \\
\cmidrule{2-11}          & Top5  & Task1 & 88.84  & 99.21  & 49.71  & 99.23  & 98.70  & 99.29  & 99.34  & \textbf{99.44 } \\
          & Acc(\%) & Task2 & 89.54  & 99.13  & 56.71  & 99.09  & 98.45  & 99.26  & 99.26  & \textbf{99.38 } \\
    \midrule
    \multicolumn{1}{l|}{ } & \multirow{2}[2]{*}{Loss} & Task1 & 2.30  & 0.87  & 2.30  & 0.87  & 2.30  & 0.70  & 0.76  & \textbf{0.49 } \\
          &       & Task2 & 2.30  & 0.93  & 2.30  & 0.93  & 2.30  & 0.78  & 0.80  & \textbf{0.47 } \\
\cmidrule{2-11}    \multicolumn{1}{c|}{Modified} & Top1  & Task1 & 9.81  & 65.32  & 9.55  & 65.47  & 16.03  & 68.06  & 66.24  & \textbf{73.10 } \\
    \multicolumn{1}{c|}{LeNet5} & Acc(\%) & Task2 & 10.87  & 64.42  & 10.77  & 64.69  & 14.97  & 67.63  & 66.03  & \textbf{72.58 } \\
\cmidrule{2-11}          & Top5  & Task1 & 51.39  & 98.58  & 51.14  & 98.61  & 64.51  & 98.82  & 98.64  & \textbf{99.10 } \\
          & Acc(\%) & Task2 & 52.05  & 98.03  & 51.80  & 98.08  & 64.14  & 98.53  & 98.34  & \textbf{99.02 } \\
    \midrule
    \multicolumn{1}{c|}{ } & \multirow{2}[2]{*}{Loss} & Task1 & 1.32  & 0.45  & 2.19  & 0.44  & 0.77  & 0.28  & 0.40  & \textbf{0.23 } \\
          &       & Task2 & 1.47  & 0.52  & 2.19  & 0.51  & 0.88  & 0.49  & 0.48  & \textbf{0.46 } \\
\cmidrule{2-11}    \multicolumn{1}{c|}{Modified} & Top1  & Task1 & 48.25  & 80.67  & 23.60  & 81.00  & 67.94  & 80.14  & 79.75  & \textbf{82.86 } \\
    \multicolumn{1}{c|}{VGG} & Acc(\%) & Task2 & 44.74  & 80.59  & 21.72  & 80.83  & 67.88  & 79.75  & 79.55  & \textbf{83.19 } \\
\cmidrule{2-11}          & Top5  & Task1 & 94.96  & 99.57  & 75.39  & 99.60  & 98.81  & 99.55  & 99.54  & \textbf{99.67 } \\
          & Acc(\%) & Task2 & 93.51  & 99.56  & 73.41  & 99.60  & 98.64  & 99.57  & 99.56  & \textbf{99.69 } \\
    \bottomrule
    \end{tabular}%
\end{center}
 \end{table}

\subsection{GML2O}
In this subsection, we evaluate our GML2O method. Firstly, we compare the DSSMG method with the SMG method, which utilizes a single stochastic gradient. This analysis allows us to highlight the advantages of employing multiple samples in the DSSMG method. Subsequently, we evaluate the performance of GML2O on CNN1, employing the DSSMG method as a backtracking step. The parameters of the learning optimizer ML2O are selected to be ${\Theta}$ in subsection \ref{sec:Generalization}.

\begin{figure}[h!]
\centering
\subfloat[BK1]{\includegraphics[scale=0.3]{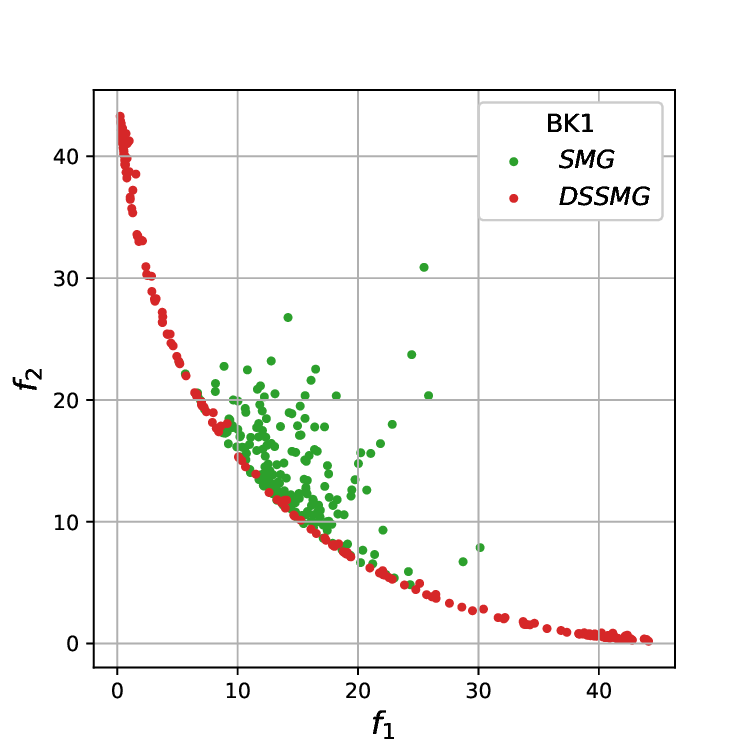}}
\subfloat[DOG1]{\includegraphics[scale=0.3]{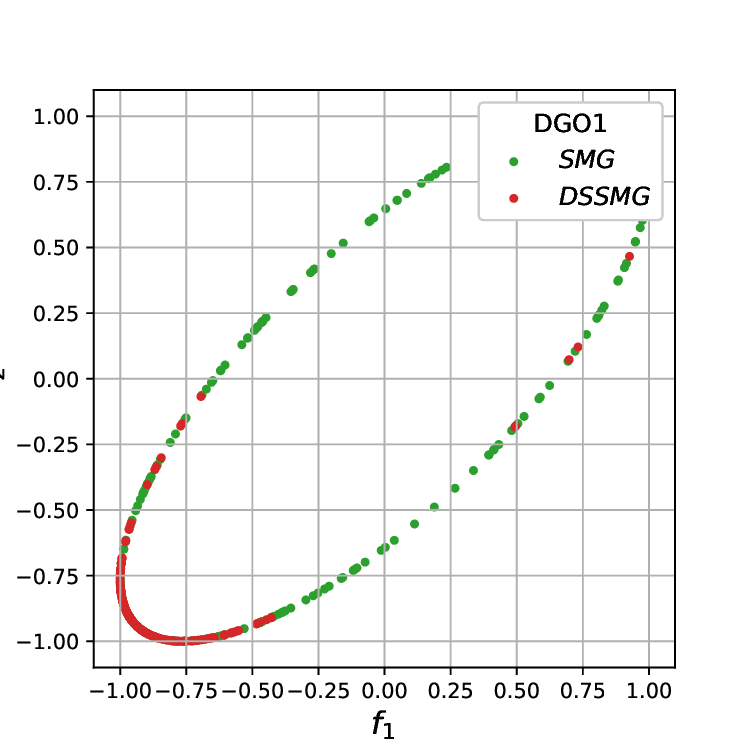}}
\subfloat[Lov1]{\includegraphics[scale=0.3]{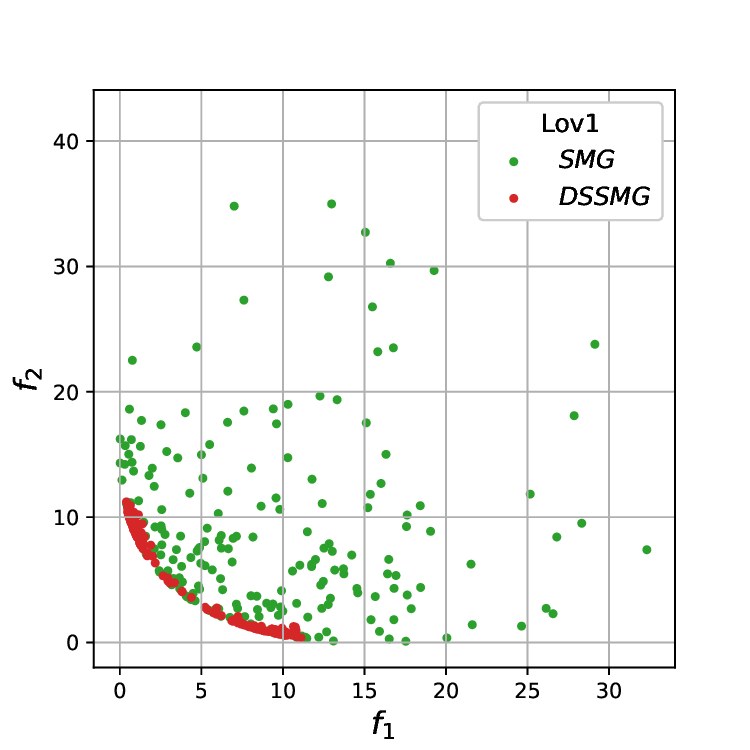}}
\subfloat[MOP5]{\includegraphics[scale=0.3]{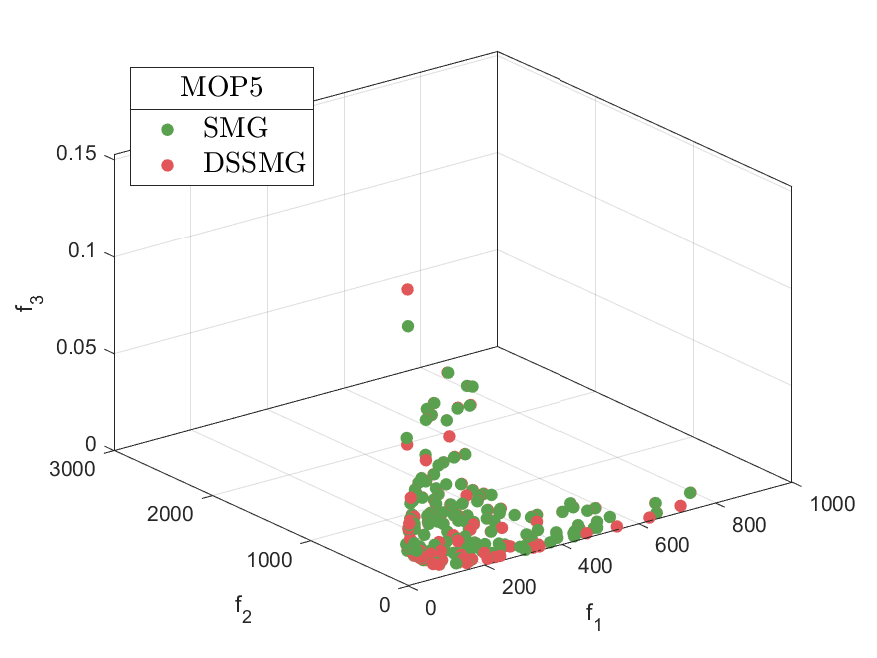}}
\caption{Comparison of single gradient and DSSMG method in portraying the traditional MOO Pareto fronts}
\label{fig:smga}
\end{figure}

\textbf{DSSMG} ~~We conducted a comparative analysis between the SMG approach and the DSSMG approach on four MOO problems. These MOO problems comprised three 2-objective functions (BK1, DOG1, Lov1) and a 3-objective function (MOP5), with initialized box constraints $[lb,ub]$ as outlined in \cite{assunccao2021conditional}. To introduce noise into the optimization process, we sampled the independent variables of each function from a normal distribution with a mean of 0 and a standard deviation of 0.1 times the range $[lb,ub]$. We set the number of iterative steps to $K=100$, the initial number of points to 200, and the optimization step size to 0.5. The dynamic sample size was determined as ${{N_k} = \max \left\{ {{N_B},{k^q}} \right\}}$, where $N_B=32$ and $q=0.1$. Figure \ref{fig:smga} demonstrate that the dynamic sample gradient approach effectively mitigates the impact of noise perturbations on the algorithm, and the DSSMG method yields a higher quality frontier surface. Specifically, for the BK1 problem, the Pareto optimal solutions identified by SMG are concentrated in the middle, whereas the frontier surface obtained by DSSMG is more comprehensive and encompasses a greater number of weakly efficient solutions.

\begin{figure}[ht]
\centering
\includegraphics[scale=0.3]{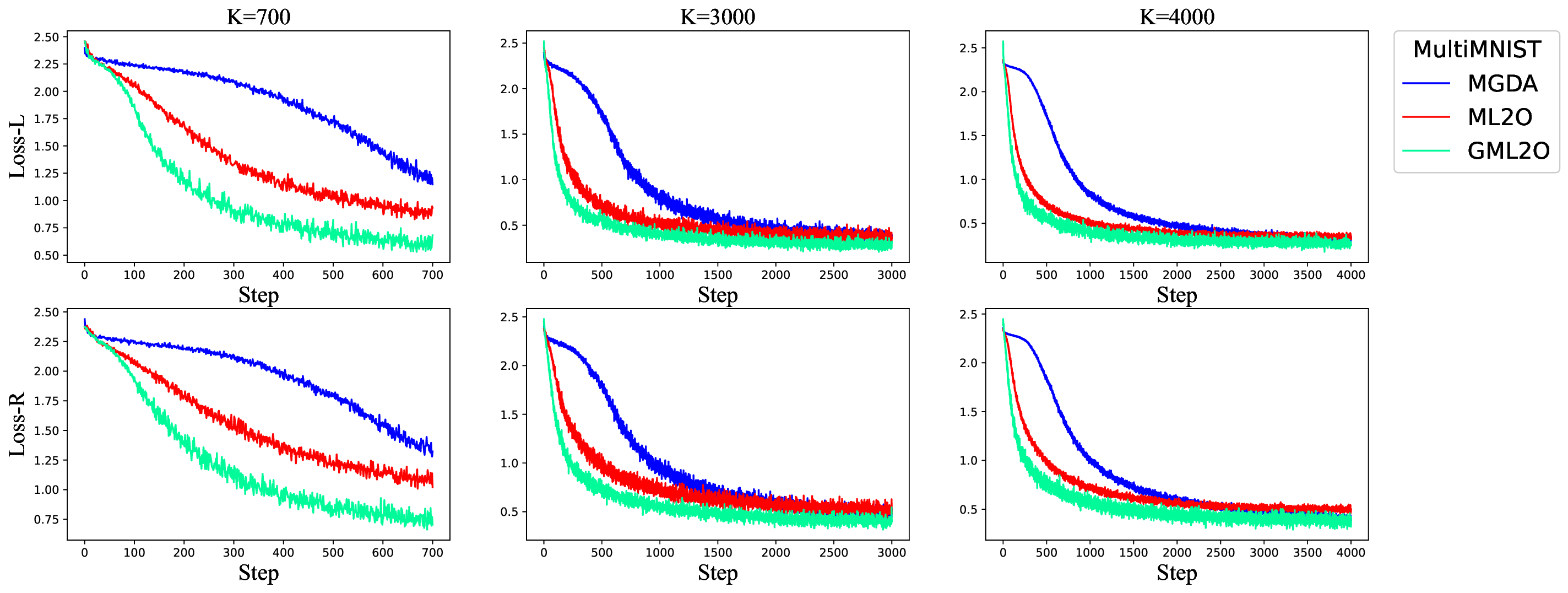}
\caption{Performance of GML2O. Comparison of MGDA, ML2O and GML2O optimized for CNN1 at K=700, K=300 and K=4000. where ML2O is selected same as in subsection \ref{sec:Generalization}. }
\label{fig:safe}
\end{figure}

\textbf{GML2O} ~~In this evaluation, we investigate the efficacy of our proposed GML2O optimizer. We utilize the DSSMG method with parameters ${N_b=1}$ and ${q=0.1}$ for backtracking update. The dataset employed for evaluation is MultiMNIST, and we conduct optimization steps in three variations, 700, 3000, and 4000 steps, respectively. Our purpose is to compare the performance of MGDA, ML2O, and GML2O as optimizers for training CNN1. The results, as depicted in Figure \ref{fig:safe}, clearly demonstrate that the loss function curve of GML2O consistently outperforms those of MGDA and ML2O. Furthermore, by incorporating the backtracking step, we observe a further enhancement in the effectiveness of ML2O, making GML2O the superior optimizer in terms of overall performance and convergence.

\begin{table}[h!]\footnotesize\tabcolsep 16pt
\begin{center}
\caption{Loss function values of MGDA, ML2O and GML2O for optimizing CNN1 at K=700, K=300 and K=4000}\label{tab:safe}\vspace{-2mm}
\end{center}
\begin{center}
    \begin{tabular}{c|c|c| ccc}
    \toprule
    \multicolumn{3}{c|}{Optimizer} & MGDA  & ML2O  & GML2O \\
    \midrule
    \multirow{2}[2]{*}{CNN1} & \multirow{2}[2]{*}{Loss} & Task1 & 1.15  & 0.93  & \textbf{0.68 } \\
          &       & Task2 & 1.32  & 1.02  & \textbf{0.70 } \\
    \midrule
    Modified & \multirow{2}[1]{*}{Loss} & Task1 & 0.39  & 0.36  & \textbf{0.32 } \\
    LeNet5 &       & Task2 & 0.53  & 0.49  & \textbf{0.40 } \\
    \midrule
    Modified & \multirow{2}[1]{*}{Loss} & Task1 & 0.33  & 0.32  & \textbf{0.25 } \\
    VGG   &       & Task2 & 0.39  & 0.49  & \textbf{0.32 } \\
    \bottomrule
    \end{tabular}%
\end{center}
 \end{table}
Table \ref{tab:safe} presents the loss function values corresponding to MGDA, ML2O and GML2O across three step settings. Consistently, GML2O exhibits the lowest loss values among all methods, corroborating the findings depicted in Figure \ref{fig:safe}. Specifically, GML2O achieves losses of $(0.68, 0.70)$, $(0.32, 0.40)$, and $(0.25, 0.32)$ for the three learners, respectively (where the first value corresponds to the classification loss of the left image and the second value corresponds to the classification loss of the right image). These values are noticeably lower than those obtained by the hand-crafted algorithm MGDA and the learning method ML2O.
\section{Conclusion}
In this paper, we proposed a learning-based method ML2O for training a neural network to solve MOO problem. This method can alleviate the shortcomings of the traditional optimization paradigm of MOO due to the strong reliance on manual design. The key strategy of our method is to consider the process of MOO as a learning problem and use LSTM to learn the update direction from multiple gradients by offline training. Furthermore, we designed a guarded version of ML2O called GML2O, which uses a guardian criterion to ensure that the updates generated by a learning method are not inferior to a certain converged method.
When the proposed DSSMG method is chosen as a fallback update for the guardian criterion, we proved that the sequence generated by GML2O converges to a Pareto critical point.
Numerical experiments on various MTL network training problems indicate that our method outperforms trade-offs and gradient-based methods in terms of losses and classification accuracy, and it is demonstrated to be generalizable under different hyperparameter settings, datasets and network architectures.

In the future, we are interested in further bridging the design methods in MOO \cite{gonccalves2022globally} with learning optimization methods to develop model-based L2O for MOO problems.
In recent years, model-based optimization has been widely studied and effectively applied to machine learning, especially deep learning \cite{shlezinger2022model, chen2022learning}.
Instead of using a generic LSTM, the iterative format of such L2O methods is designed by analytical optimization algorithms.
Unlike the model-free learning optimizer discussed in this paper, which requires a large number of training samples to search for a well-performing learning optimizer ``from scratch" \cite{chen2017learning}, the model-based L2O takes the existing optimization methods as the starting point for learning, reducing the search space for the learning optimizer.
Therefore, given that model-based learning optimizer overcomes the shortcomings that lacks convergence guarantees and high demand for training samples \cite{chen2017learning}, it is worthwhile for us to explore the potential of model-based L2O in addressing MOO optimization problems.
\subsubsection*{Acknowledgments}
This work was funded by the Major Program of the National Natural Science Foundation of China (Grant Nos. 11991020, 11991024); the National Natural Science Foundation of China (Grant Nos. 11971084, 12171060); NSFC-RGC (Hong Kong) Joint Research Program  (Grant No. 12261160365); the Team Project of Innovation Leading Talent in Chongqing (Grant No. CQYC20210309536); the Natural Science Foundation of Chongqing of China (Grant No. ncamc2022-msxm01), the Major Project of Science and Technology Research Rrogram of Chongqing Education Commission of China (Grant No. KJZD-M202300504) and the Foundation of Chongqing Normal University (Grant Nos. 22XLB005, 22XLB006).

\end{document}